\documentclass{article}
\usepackage[margin=1in]{geometry} 
\usepackage[round]{natbib}

\usepackage{graphicx} 
\usepackage{xcolor}
\usepackage{subcaption}
\usepackage{float}
\usepackage{placeins}
\usepackage[most]{tcolorbox}
\usepackage[linesnumbered,ruled,vlined]{algorithm2e}
\usepackage{titletoc}
\usepackage{longtable}
\usepackage{bm}
\usepackage{tabularx}
\usepackage{pdflscape}

\definecolor{mitred}{RGB}{163,31,52}
\definecolor{DarkGreen}{rgb}{0.1,0.5,0.1}
\definecolor{DarkRed}{rgb}{0.5,0.1,0.1}
\definecolor{DarkBlue}{rgb}{0.1,0.1,0.5}
\definecolor{Gray}{rgb}{0.2,0.2,0.2}
\usepackage[pdftex]{hyperref}
\hypersetup{
    unicode=false,          
    pdftoolbar=true,        
    pdfmenubar=true,        
    pdffitwindow=false,      
    pdfnewwindow=true,      
    colorlinks=true,       
    linkcolor=DarkBlue,          
    citecolor=DarkBlue,        
    filecolor=DarkRed,      
    urlcolor=DarkBlue,          
    pdftitle={Aggregation Hides Out-of-Distribution Generalization Failures from Spurious Correlations},
    pdfauthor={Olawale Salaudeen},
}

\usepackage{definitions}

\newcommand{\condensedVisionModelList}{\texttt{AlexNet}, \texttt{ResNet-(18/34/50/101/152)}, \texttt{DenseNet-(121/161/169/201)}, \texttt{MobileNet (V2/V3 Small/V3 Large)}, \texttt{EfficientNet-(B0/B1/B3/B7)}, \texttt{ConvNeXt-(Tiny/Small/Base/Large)}, \texttt{ViT-(B/16, B/32, L/16)}, \texttt{Swin Transformer-(Tiny/Small/Base)}, \texttt{RegNet-Y (400MF/800MF/1.6GF/3.2GF/8GF)}, \texttt{VGG-(11/13/16/19)}, \texttt{SqueezeNet (1.0/1.1)}, and \texttt{Inception v3}.}

\newcommand{\condensedTextModelList}{%
\texttt{BERT-(base/large)}, \texttt{SciBERT}, \texttt{RoBERTa-(base/large)}, \texttt{BioBERT}, \texttt{LegalBERT}, \texttt{FinBERT}, 
\texttt{ALBERT-(v1/v2, base/large/xlarge/xxlarge)}, 
\texttt{DeBERTa-v2-(xsmall/small/base/large)}, 
\texttt{Longformer-(base/large)}, 
\texttt{DistilBERT-(base/cased, distilled)}, 
\texttt{T5-(small)}, 
\texttt{BART-(base/large/mnli)}, and
\texttt{GPT-2-(small)}. 
}

\newtheorem{definition}{Definition}
\newtheorem{lemma}{Lemma}

\newtheorem{proposition}{Proposition}

\usepackage[utf8]{inputenc} 
\usepackage[T1]{fontenc}    
\usepackage{hyperref}       
\usepackage{url}            
\usepackage{booktabs}       
\usepackage{amsfonts}       
\usepackage{nicefrac}       
\usepackage{microtype}      
\usepackage{xcolor}         
\usepackage{makecell}

\title{Aggregation Hides Out-of-Distribution Generalization Failures from Spurious Correlations}
\author{%
\textbf{Olawale Salaudeen} \quad
\textbf{Haoran Zhang} \quad
\textbf{Kumail Alhamoud} \\
\textbf{Sara Beery} \quad
\textbf{Marzyeh Ghassemi} \\
Massachusetts Institute of Technology\\
Correspondence to \texttt{olawale@mit.edu}.
}

\begin{document}

\maketitle

\begin{abstract}
Benchmarks for out‑of‑distribution (OOD) generalization frequently show a strong positive correlation between in‑distribution (ID) and OOD accuracy across models, termed ``accuracy‑on‑the‑line.'' This pattern is often taken to imply that spurious correlations---correlations that improve ID but reduce OOD performance---are rare in practice. We find that this positive correlation is often an artifact of aggregating heterogeneous OOD examples. Using a simple gradient‑based method, \texttt{OODSelect}, we identify semantically coherent OOD subsets where accuracy on the line does not hold. {\bf Across widely used distribution shift benchmarks, the \texttt{OODSelect} uncovers subsets, sometimes over half of the standard OOD set, where higher ID accuracy predicts lower OOD accuracy.} Our findings indicate that aggregate metrics can obscure important failure modes of OOD robustness. We release code and the identified subsets to facilitate further research.
\end{abstract}

\section{Introduction}
Benchmarks for out-of-distribution (OOD) generalization have shown a consistent pattern that models performing well on the training distribution also perform well out-of-distribution, a trend known as \emph{accuracy-on-the-line} (AoTL)~\citep{miller2021accuracy, taori2020measuring}. This pattern has often been interpreted as evidence that spurious correlations---features that improve in-distribution (ID) accuracy but harm OOD performance---are uncommon in practice. We show that this apparent robustness is misleading. When OOD data are disaggregated, large and semantically coherent subsets emerge where higher ID accuracy predicts lower OOD accuracy, a phenomenon we term \emph{accuracy-on-the-inverse-line} (AoTIL). These hidden subsets reveal that aggregation can mask major failures of OOD robustness, suggesting that existing benchmarks may underestimate the prevalence and impact of spurious correlations.

The promise of machine learning lies in generalization, the ability to perform a task on new data with similar effectiveness as on the training data~\citep{blumer1989learnability, vapnik1999nature, shalev2014understanding, zhang2016understanding, belkin2019reconciling}. Yet models deployed in a dynamic world often encounter data from different distributions~\citep{shimodaira2000improving, moreno2012unifying} and fail. For instance, a medical diagnosis model trained on data from one hospital may perform poorly in another with distinct demographics or equipment~\citep{zech2018variable, yang2024limits}, and an animal classifier may misclassify images captured under new conditions~\citep{beery2018recognition, xiao2020noise}. Generalization under such shifts, from in-distribution (ID) training to out-of-distribution (OOD) testing or deployment, defines domain generalization~\citep{zhou2022domain, wang2022generalizing}.

These observations motivate a closer examination of what benchmark correlations actually reveal about robustness and when they conceal spurious mechanisms that undermine OOD generalization.

\begin{figure}[t!]
    \centering
    \begin{subfigure}[t]{0.48\textwidth}
        \includegraphics[width=\linewidth]{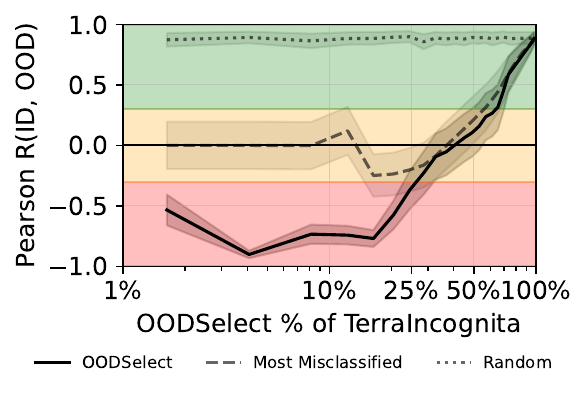}  
        \caption{\textbf{AoTIL:} With original OOD data, ID and OOD accuracy across a set of models has a Pearson correlation of $0.89$. \texttt{OODSelect} finds up to $1000$ ($\sim16\%$) examples from \texttt{L46} on the same models with an ID-OOD correlation of $-0.77$.} 
    \end{subfigure}
    \hfill
    \begin{subfigure}[t]{0.48\textwidth}
        \includegraphics[width=\linewidth]{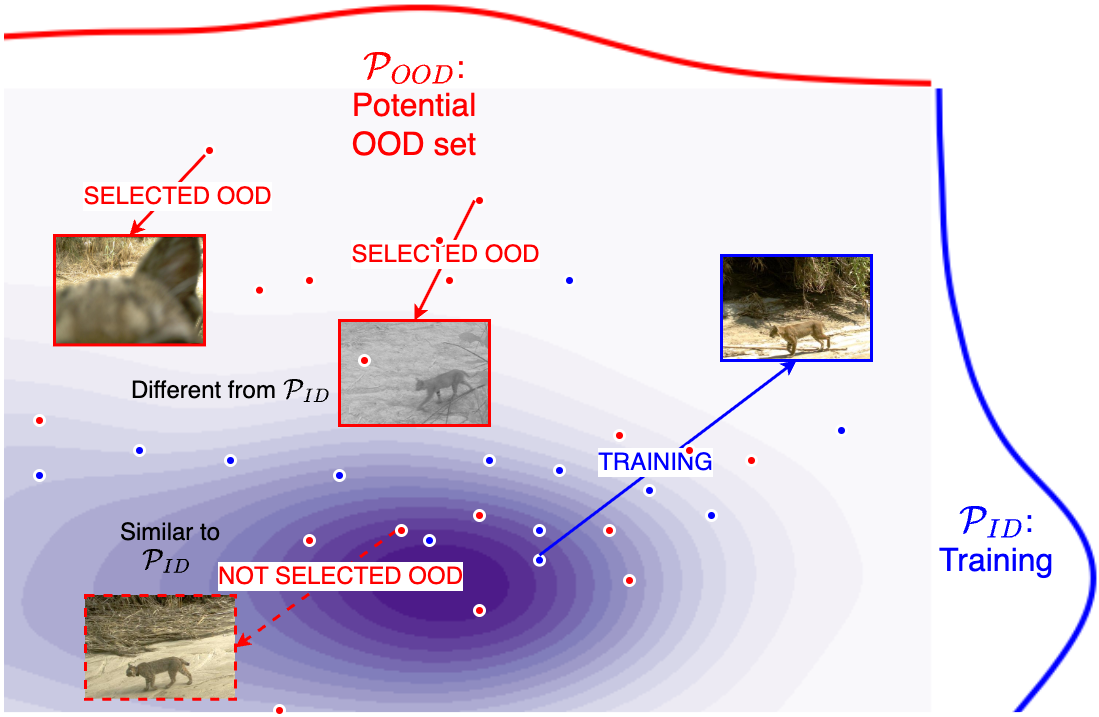}
        \caption{\textbf{\texttt{OODSelect} Strategy:} Excluded examples resemble the training distribution (e.g., centered bobcats in daylight), while included \texttt{OODSelect} examples differ (e.g., occluded bobcats, infrared camera capture).}
    \end{subfigure}
    \caption{\textbf{Aggregation Masking AoTIL.} Consider Terra Incognita, where ID data are drawn from camera traps at locations \texttt{L100, L38, L43}, and OOD data from \texttt{L46}~\citep{beery2018recognition}. Aggregation masks the effect of spurious correlations on generalization, such as daylight, even though a substantial number of OOD samples are still systematically misclassified. Note that \texttt{OODSelect} examples differ from the most misclassified examples, which always have an ID-OOD accuracy correlation of near zero. Confidence intervals correspond to 95\% Fisher z-intervals.}
    \label{fig:summary_figure}
\end{figure}

In this work, we establish the existence of large and semantically coherent OOD subsets in state-of-the-art datasets with accuracy on the inverse line. Specifically, {\bf our contributions} are:
\begin{itemize}
    \item We show that in state-of-the-art domain generalization benchmarks, there exist large, semantically meaningful OOD subsets—sometimes up to over half of the data—with correlations low as $-0.9$ Pearson $R$ (Figure~\ref{fig:corr_summary}). The familiar {\em accuracy-on-the-line} trend only emerges once such subsets are aggregated with the rest of the data.
    
    \item We show that these subsets are not arbitrary: for example, in Chest X-ray diagnosis tasks, models that improved overall performance performed \emph{worse} on patients with pleural conditions and enlarged cardiomediastinum. 
    
    \item We propose \texttt{OODSelect}, a simple yet effective selection procedure to identify such subsets across datasets, when they exist.
    
    \item We provide the identified subsets for state-of-the-art datasets, including those from DomainBed~\citep{gulrajani2020search} and WILDS~\citep{koh2021wilds}, to facilitate future research (included in the supplementary material).

\end{itemize}

We provide the \href{https://github.com/olawalesalaudeen/OODSELECT}{code and selected subsets}\footnote{https://github.com/olawalesalaudeen/OODSELECT} for our proposed OOD selection method and analysis.

\section{Background and Related Work}
The field of OOD generalization aims to develop models that are robust to spurious correlations~\citep{zhou2022domain, wang2022generalizing}. Many of the state-of-the-art methods in domain generalization rely on notions of distributional invariance~\citep{arjovsky2019invariant, krueger2021out}; often using causal motivations~\citep{peters2016causal, heinze2018invariant, salaudeen2024causally, salaudeen2024domain}.
Progress in the field of domain generalization has primarily been evaluated by two benchmark suites: DomainBed~\citep{gulrajani2020search} and WILDS~\citep{koh2021wilds}. However, various studies have suggested that none of the proposed domain generalization methods consistently outperform naive empirical risk minimization on these benchmarks~\citep{gulrajani2020search, koh2021wilds, yang2023change}.
Moreover, previous work has suggested that improving ID accuracy tends to improve OOD accuracy, i.e., a strong correlation between ID and OOD accuracy holds, termed {\em accuracy on the line}~\citep{miller2021accuracy, taori2020measuring, saxena2024predicting, sanyal2024accuracy}. However,~\citet{NEURIPS2023_e304d374} demonstrate that with a more diverse selection of models, a fraction of real-world datasets do indeed exhibit other correlations besides strong and positive correlations between ID and OOD accuracy. Furthermore,~\citet{salaudeen2025domaingeneralizationbenchmarksaccuracy} provides a theoretical analysis that suggests prioritizing datasets without accuracy on the line; our proposed method provides OOD sets that satisfy such conditions by selecting subsets of existing benchmarks with accuracy on the line.

Existing subset discovery methods—such as \texttt{Slice Finder}, \texttt{SSD++}, and \texttt{DivExplorer}\citep{polyzotis2019slice,proencca2022robust,pastor2021identifying,subbaswamy2020development}—rely on explicit grouping cues, categorical features, or annotated attributes to define candidate subsets. In contrast, our setting assumes no access to such metadata and requires model-agnosticism, motivating a simple yet effective selection approach. \texttt{Influence functions}~\citep{koh2017understanding} may appear suitable at first glance, but they rank {\em training} points by leave-one-out influence, rather than partitioning the {\em test/OOD} set. Thus, applying influence functions in this context would still require an additional heuristic to define coherent subsets, while also inheriting known fragilities in modern deep networks~\citep{basu2020influence,epifano2023revisiting,bae2022if,grosse2023studying,koh2019accuracy,hu2024most}.

Our proposed method in the next section provides a simple, efficient, yet effective approach.

\section{Methodology}
First, we define the correlation property that is used to determine AoTL or AoTIL.

\begin{definition}[Correlation Property;~\cite{miller2021accuracy}]\label{def:corr_prop}
Define $a \in \RR$, $\epsilon \ge 0$, and $\Phi^{-1}$ as the inverse Gaussian cumulative density function. The {\em correlation property} is defined as 
\begin{equation}
\left|\Phi^{-1}\left(\acc_{P_\ID}(f)\right) - a \cdot \Phi^{-1}\left(\acc_{P_\OOD}(f)\right)\right| \le \epsilon \quad \forall f.
\end{equation}
\end{definition}


Definition~\ref{def:corr_prop} implies:
\begin{equation}
|\text{Pearson}~R(X, Y)| \gtrsim 1 - \frac{\epsilon}{|a| \cdot \sigma_Y},
\end{equation}
where $\sigma_Y$ is the standard deviation of $Y$. Thus, the correlation property implies that the transformed ID and OOD accuracies lie approximately on a line and are strongly linearly correlated. Moreover, the sign of the Pearson correlation is determined by the sign of $a$: if $a > 0$, the correlation is positive, and if $a < 0$, the correlation is negative.

\eparagraph{Problem Setup.}
Suppose we have $N$ models $f_i$ and $d$ potential OOD examples. Let $\Zb \in \RR^{N \times d}$ where $\Zb_{ij}$ is $1$ if model $f_i$ correctly classifies example $j$ and $0$ otherwise. Define $\accidb \in \RR^N$ where $(\accidb)_i$ is the held-out in-distribution accuracy of model $f_i$.  In this work, we are always operating on $0-1$-clipped probit transform of accuracy. Define a sample selection vector $\sbb \in \{0, 1\}^d$ that indicates which examples to select from the candidate OOD set, and denote the selected OOD accuracy for model $f_i$

\begin{minipage}[t]{0.3\textwidth}
\centering
\begin{equation}
(\accoodsb)_i = \frac{\Zb_{[i,:]} \sbb}{\|\sbb\|_1} 
\end{equation}
\end{minipage}
\hfill
\begin{minipage}[t]{0.65\textwidth}
\centering
\begin{equation}\label{eq:corr}
\text{ and \quad }
\text{corr}({\accidb}, {\accoodsb}) = 
\frac{{\accidb}^\top {\accoodsb}}{\sqrt{\|{\accidb}\|^2\|{\accoodsb}\|^2}},
\end{equation}
\end{minipage}

where $\text{corr}$ is the Pearson correlation between ID and OOD accuracies. Note that the probit-transformed accuracies are mean-centered before computing the correlations.

\SetKwInput{KwIn}{Input}
\SetKwInput{KwOut}{Output}
\SetNlSty{textbf}{}{:}
\SetAlgoNlRelativeSize{-1}
\DontPrintSemicolon

\begin{algorithm}[t]
\caption{\texttt{OODSelect}: Selecting OOD subsets without accuracy-on-the-line}
\label{algo:selection}

\KwIn{
    $\Dcal_{\ID}^{\text{train}}, \Dcal_{\ID}^{\text{test}}$: in-distribution train/test splits;\\
    $\Dcal_{\OOD}$: out-of-distribution dataset;\\
    $S \in \mathbb{N}_{\le|\Dcal_{\OOD}|}$: number of OOD samples to select
}
\KwOut{Subset $\Dcal_{\OOD}^\sbb \subset \Dcal_{\OOD}$ of size $S$}
\vspace{0.5em}

\SetKwProg{Fn}{Procedure}{}{}

Train $N_{\text{models}}$ diverse models on $\Dcal_{\ID}^{\text{train}}$.\;

Let ${\accidb} \in \mathbb{R}^{N_{\text{models}}}$ be the vector of probit-transformed accuracies, where ${\accidb}_i$ denotes the accuracy of model $i$ on $\Dcal_{\ID}^{\text{test}}$.\;
 
Construct binary matrix $\Zb \in \{0,1\}^{N_{\text{models}} \times |\Dcal_{\OOD}|}$ where:\;
\Indp
\[
\Zb_{ij} = 
\begin{cases}
1 & \text{if model } i \text{ correctly classifies OOD sample } j \\
0 & \text{otherwise}
\end{cases}.
\]
\Indm

Let ${\accoodsb} \in \mathbb{R}^{N_{\text{models}}}$ denote the per-model average accuracy vector across the OOD examples selected by $\sbb$; that is, $${\accoodsb} = \frac{1}{\|\sbb\|_1} \Zb_{\sbb},$$ where $\Zb_{\sbb}$ denotes the columns of $\Zb$ indexed by $\sbb$.\;

Solve the optimization in Equation~\ref{eq:main_objective} to find $\Dcal_{\OOD}^\sbb$ from $\Dcal_{\OOD}$.\;

\end{algorithm}

\eparagraph{Objective.}
We aim to learn a selection vector $\sbb\in\{0,1\}^d$ (with $S = \|\sbb\|_1$ fixed) to minimize the correlation between ${\accidb}$ and ${\accoodsb}$---ideally with large $S$.

Importantly, a subset achieving weak or negative correlation may not exist, particularly if there are no spurious correlations with respect to the ID and OOD distributions. Additionally, the change in Pearson $R$ from adding a model or OOD example is bounded by $\mathcal{O}(C/\sqrt{m})$, where $m$ is the number of models or OOD examples and $C$ depends on the accuracy range and the Lipschitzness of $\Phi^{-1}$ (theoretical analysis provided in Appendix~\ref{sec:theory} Lemma~\ref{lem:bnd_model}-\ref{lem:bnd_example}).

We consider a constrained objective for selecting $S$ OOD examples:
\begin{align}
    \min_{\sbb \in \{0,1\}^d} \quad \text{corr}({\accidb}, {\accoodsb}) \nonumber \quad 
    \text{subject to} \quad \|\sbb\|_1 = S.
\end{align}
We relax this objective to:
\begin{align} \label{eq:main_objective}
    \min_{\sbb \in [0,1]^d} \quad &\text{corr}({\accidb}, {\accoodsb}) + \lambda \cdot \left(S - \|\sbb\|_1\right)^2,
\end{align}
where $\sbb$ is the output of a sigmoid function in practice.

\eparagraph{Soundness of the relaxation and optimization.}
Our objective is non-convex and non-submodular (Proposition~\ref{prop:nonsubmodularity}), but Lipschitz-continuous (Lemma~\ref{lem:pearson_lipschitz}). While global optimization is intractable, the Lipschitz property ensures stable gradients and bounded progress under descent, enabling convergence toward near-binary stationary points that approximate the discrete optima. Non-submodularity also eliminates greedy selection as an optimal strategy. We use the Adam optimizer~\citep{kingma2014adam} to optimize Equation~\ref{eq:main_objective}. We use a cosine annealing schedule to adjust the learning rate and  $\lambda$~\citep{loshchilov2016sgdr}. Additional details are available in Appendix~\ref{sec:empirical_details}.

\eparagraph{On Selected Subsets.} Although we are free to choose $S$ examples, a subset that makes the ID-OOD Pearson $R$ negative is not guaranteed to exist. The OOD accuracy of each model is an average over the selected examples. The subset must systematically up- or down-weight groups of examples on which higher-accuracy ID models tend to underperform relative to lower-accuracy ID models. We provide evidence that finding a large sign-flipping subset is evidence of latent structure or spurious shortcuts in the data, not a trivial consequence of sub-sampling. Importantly, we do not select models or alter the ID accuracies; we always correlate the same length-$N$ vectors, only the OOD accuracy values change through the choice of examples.

For brevity, we reserve details of other theoretical analyses for Appendix~\ref{sec:theory}, as our results are included for thoroughness, but they are standard~\citep{bertsekas1997nonlinear, nocedal1999numerical}.

\newcommand{\corr}{\text{corr}}

\eparagraph{Fisher Confidence Intervals.} In our estimate of correlations, we compute Fisher $z$ intervals for each correlation estimate, indicating the range of variability expected from estimation; overlapping bars suggest that differences could be arbitrary, while non-overlapping drops signal meaningful differences in correlation.

\section{Experiments} \label{sec:empirical_results}
\begin{table}[t!]
  \centering
  \caption{\textbf{Dataset Summary.} Each dataset defines a classification task across multiple domains. Full OOD Size refers to the size of the OOD dataset.  We select the full dataset to apply \texttt{OODSelect} according to splits in DomainBed~\citep{gulrajani2020search} and WILDS~\citep{koh2021wilds}. WILDSCamelyon-H4/H5 refer to the versions of the dataset where hospitals 4 and 5 are considered OOD distinctly. Importantly, we never train models on either and separate them here because they have distinct properties and results. WILDSCivilComments is a text dataset.}
  \resizebox{\textwidth}{!}{%
  \begin{tabular}{lccccc}
    \toprule
    \bfseries Dataset & \bfseries Task (\# classes) & \bfseries Domains (\#) & \bfseries \makecell{Full OOD\\Size} & \bfseries\makecell{Largest \texttt{OODSelect}\\Subset w/ AoTIL ($-0.3$)} & \bfseries \# Models\\
    \midrule
    \makecell[l]{\texttt{Chest Xrays}} & \makecell[c]{Finding vs.\\ No Finding} (2) & \makecell[c]{Hospital\\systems} (5) & 71433 & 55000 (75\%) & 1800 \\
    \makecell[l]{\texttt{PACS}} & \makecell[c]{Object\\classification} (7) & \makecell[c]{Styles} (4) & 3929 & 250 (6\%) & 2804 \\
    \makecell[l]{\texttt{VLCS}} & \makecell[c]{Object\\classification} (5) & \makecell[c]{Visual\\domains} (4) & 2656 & 800 (30\%) & 4200 \\
    \makecell[l]{\texttt{TerraIncognita}} & \makecell[c]{Wildlife\\classification} (10) & \makecell[c]{Camera\\traps} (4) & 6122 & 1500 (25\%) & 2980 \\
    \makecell[l]{\texttt{WILDSCamelyon-H4}} & \makecell[c]{Tumor vs.\\Normal} (2) & Hospitals (5) & 129838 & 35000 (25\%) & 944 \\
   \makecell[l]{ \texttt{WILDSCamelyon-H5}} & \makecell[c]{Tumor vs.\\Normal} (2) & Hospitals (5) & 146722 & 60000 (40\%) & 944 \\
    \texttt{WILDSCivilComments} & \makecell[c]{Toxic vs.\\Not Toxic} (2) & Demographics (8) & 52823 & 25000 (50\%) & 710 \\
    \bottomrule
  \end{tabular}}
  \label{tab:setup}
\end{table}
\begin{figure}[t!]
    \centering
    \includegraphics[width=\textwidth]{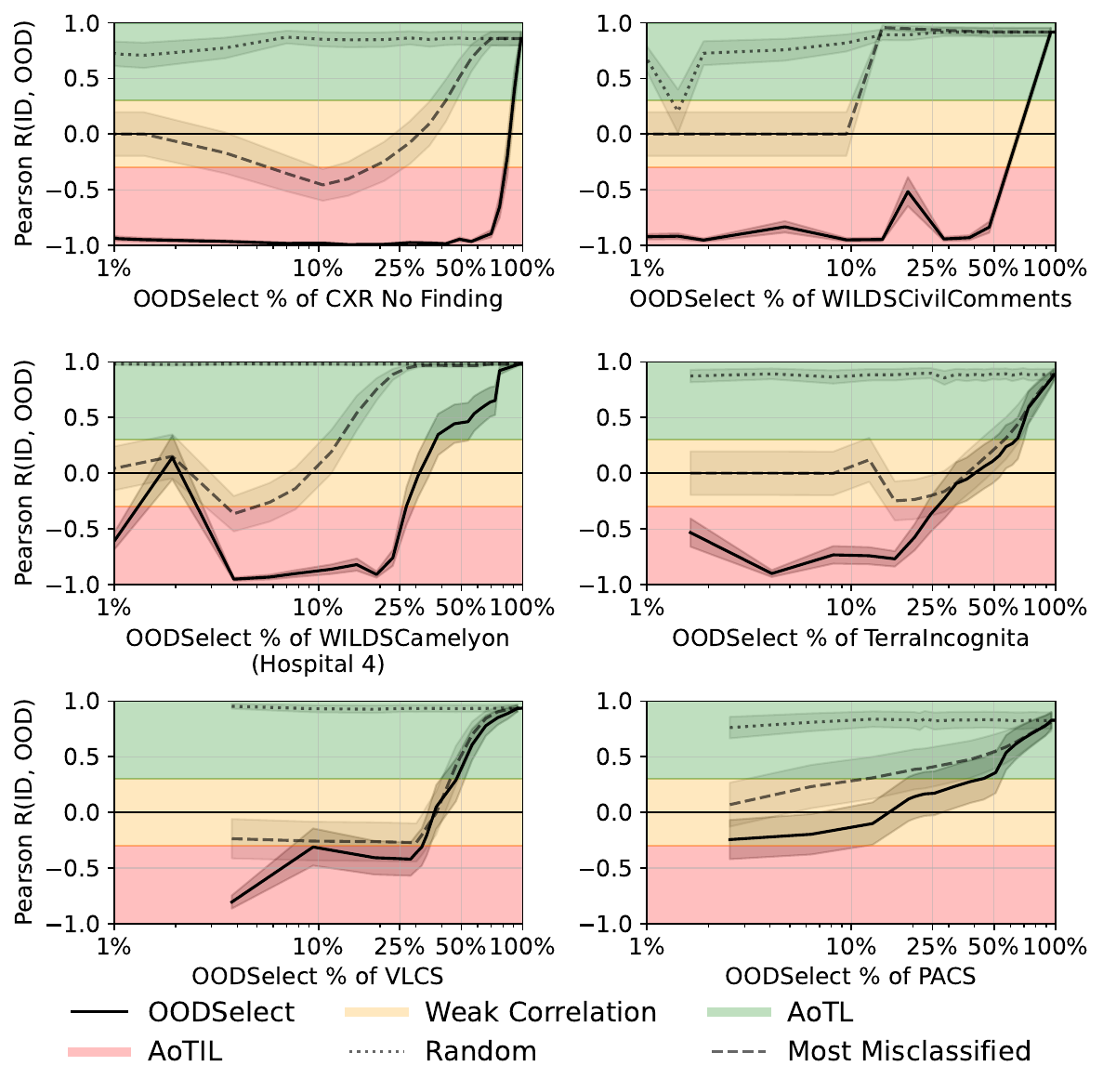}
    \caption{\textbf{Comparing AoTL and AoTIL.} Pearson Correlation between ID and OOD accuracy as a function of the number of selected OOD samples. Correlation values above 0.3 indicate AoTL, while below -0.3 is AoTIL---correlations in between are considered weak. We compare a Random Selection of data samples and the Most Misclassified at fixed size intervals from 100 to over 100,000 (normalized to sample size in the figure). Random selections yield strong positive correlation, while misclassified samples have weak correlations; that is, our method does not conflate spurious correlations with general difficulty (e.g., label noise). \texttt{OODSelect} identifies subsets where ID and OOD accuracy are negatively correlated---in one case (CXR) for over 70\% of the usual OOD dataset. This behavior is dataset-dependent due to differences in distributional properties. Table~\ref{tab:corrs} enumerates detailed correlations.}
    \label{fig:corr_summary}
\end{figure}
\eparagraph{Procedure.}
Table~\ref{tab:setup} summarizes the datasets we study. Given a typical distribution shift benchmark with at least two domains, i.e., $\Dcal = \{\Dcal_1, \Dcal_2, \ldots\}$, we fix a $\Dcal_\ID,\,\Dcal_\OOD \subset \Dcal$ pair, which are disjoint sets (concatenated) of domains. This pair denotes an experimental setting. In this work, we focus on the standard $\Dcal_\ID,\,\Dcal_\OOD$ splits the community uses for each dataset~\citep{gulrajani2020search, koh2021wilds}. For each split, we apply our methodology to identify subsets $\Dcal_\OOD^\sbb$ with AoTIL---Appendix~\ref{sec:empirical_details} Algorithm~\ref{algo:selection}.

\eparagraph{Datasets.} We consider real-world tasks and distributions such as predicting ``Finding''/``No Finding'' from {\bf Chest X-rays} where domains ID domains are from \texttt{CheXpert} (v1.0-small)~\citep{irvin2019chexpert}, \texttt{ChestXray8}~\citep{wang2017chestx}, \texttt{PadChest}~\citep{bustos2020padchest}, and \texttt{VinDr-CXR}~\citep{nguyen2022vindr}. The OOD domain is \texttt{MIMIC-CXR-JPG}~\citep{johnson2019mimic}.
We also study WILDS~\citep{koh2021wilds} benchmarks that capture real-world shifts.
\textbf{WILDS-Camelyon}~\citep{bandi2018detection} targets cancer detection from histopathology slides across hospitals.
\textbf{WILDS-CivilComments}~\citep{borkan2019nuanced, koh2021wilds} classifies online comments as toxic or non-toxic across demographic subgroups, with OOD domains defined by shifts in identity attributes such as gender, religion, and race.
We also study DomainBed~\citep{gulrajani2020search} benchmarks reflecting different forms of distribution shift: style, dataset collection, and environment. \textbf{PACS}~\citep{li2017deeper} involves object classification across artistic styles (7 classes across \textit{Photo}, \textit{Art Painting}, \texttt{Cartoon}, and \texttt{Sketch}), with \texttt{Sketch} as OOD.
\textbf{VLCS}~\citep{fang2013unbiased} spans 5 classes across 4 datasets (\texttt{VOC2007}~\citep{everingham2010pascal}, \texttt{LabelMe}~\citep{russell2008labelme}, \texttt{Caltech101}~\citep{fei2004learning}, and \texttt{SUN09}~\citep{choi2010exploiting}), capturing collection biases; \texttt{LabelMe} is OOD.
\textbf{Terra Incognita}~\citep{beery2018recognition} focuses on wildlife recognition across 4 geographic locations, with \texttt{L46} as OOD.

\eparagraph{Models.} We construct a diverse population of models by varying architecture (from VGG to Vision Transformers, listed below), pretraining weights~\citep{torchvision2016, deng2009imagenet, he2019rethinking}, initialization (from scratch and transfer learning), and hyperparameters. We train up to 4200 models (Figure~\ref{fig:num_models}) with various vision architectures, including variants of ResNets~\citep{he2016deep}, DenseNets~\citep{huang2017densely}, MobileNets~\citep{howard2017mobilenets}, ViT~\citep{dosovitskiy2020image}, VGG~\citep{simonyan2014very}, and Inception~\citep{szegedy2015going}. We do the same for our language experiments, from BERT~\citep{devlin2019bert} to GPT-2~\citep{radford2019language}. A full list of models is provided in Appendix~\ref{sec:empirical_details}.

Models are split into disjoint train, validation, and test subsets, i.e., the models used for learning the selection, cross-validation, and final testing are non-overlapping. For a given ID/OOD setting, all models are trained on the same ID training data and evaluated on a held-out ID test set and candidate OOD subsets. The resulting paired ID/OOD accuracies are used to estimate the correlation between ID and OOD performance. Further discussion on implementation is provided in Appendix~\ref{sec:empirical_details}.

\eparagraph{On the Necessary Quantity of Models.} We determine the minimum number of models to sample by thresholding the relative change in ID and OOD accuracy correlation across the full dataset. We select at least a number of models such that adding a new model changes the correlation by less than 1\%~\citep{schonbrodt2013sample, bonett2000sample}. Notably, the diversity and quantity of models we consider are orders of magnitude higher than in previous work~\citep{miller2021accuracy}; in some cases, tens vs. thousands (ours). This number is also dataset dependent; for instance, 1010 models are needed to satisfy this criterion for the VLCS dataset, while only 610 are needed for WILDS Camelyon. Further details are provided in Appendix~\ref{sec:empirical_details} Figure~\ref{fig:num_models}.

\section{Empirical Results and Discussion}

\begin{figure}[t!]
    \centering
    \begin{subfigure}[t]{0.48\textwidth}
        \centering
        \includegraphics[width=\linewidth]{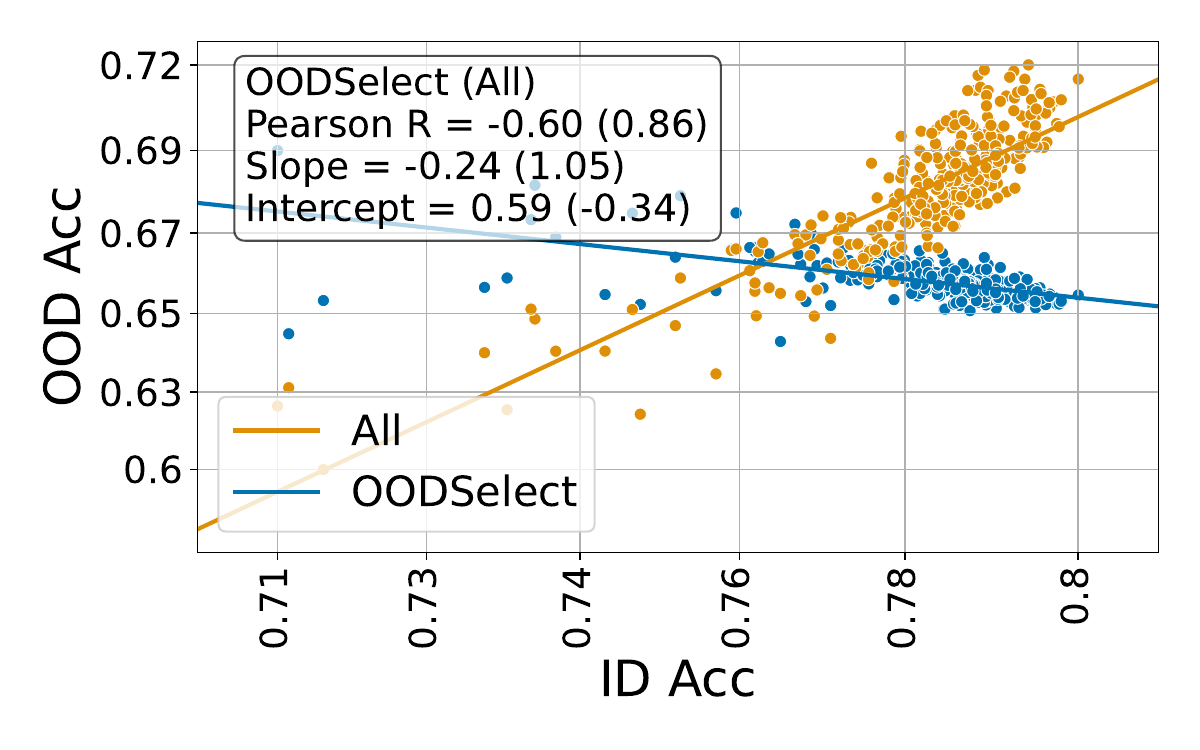}
        \caption{CXR No Finding accuracy correlations across models has a strong global correlation ($0.86$) while the OODSelect subset (55000 examples and 77\% of the full dataset) has a negative correlation ($-0.60$).}
        \label{fig:cxr_all_sub}
    \end{subfigure}
    \hfill
    \begin{subfigure}[t]{0.48\textwidth}
        \centering
        \includegraphics[width=\linewidth]{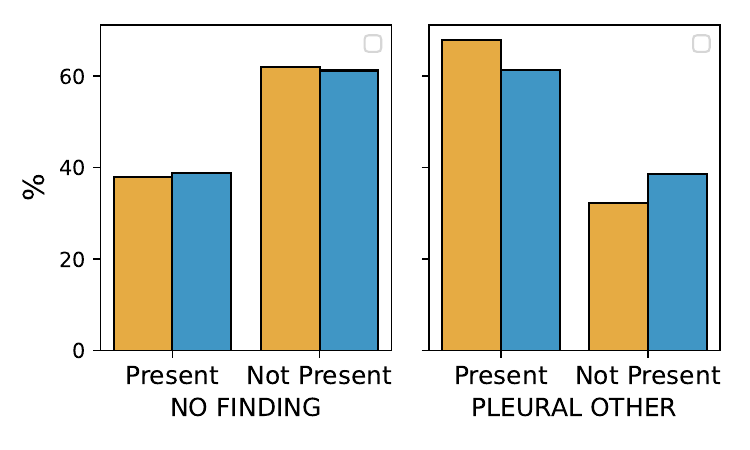}
        \caption{There is minimal label shift in the full OOD set and the OODSelect subset. No Finding has a prevalence of 39\% in the OODSelect samples and 38\% in the full dataset. However, the prevalence of ``Pleural Other'' in the full OOD set and OODSelect subset is 61\% and 68\%, respectively.}
       \label{fig:cxr_selected_sub}
    \end{subfigure}\\
    \caption{\textbf{CXR No Finding.}  Figure~\ref{fig:cxr_all_sub} suggests that poor generalization may arise for a large subset of the OOD population from reliance on spurious correlations. However, aggregation hides this failure mode since the correlation for the full OOD set is strongly positive. This selected subset also has a prevalence shift from the full dataset; statistical significance for the prevalence shift was assessed using bootstrapping with 1000 resamples.}

    \label{fig:cxr_selected_aotl}
\end{figure}

\eparagraph{Findings.} Overall, we find that many benchmarks contain \texttt{OODSelect} subsets of examples that exhibit AoTIL or a weak correlation, though the size of such subsets varies. The same benchmarks exhibit AoTL when all OOD samples are aggregated (Figure~\ref{fig:corr_summary}). 

We treat $|R| < 0.3$ as a weak Pearson correlation between ID accuracies and OOD accuracies across models. Using \texttt{OODSelect}, our method for selecting the OOD data, we uncover large variance in correlations that are hidden in the full splits. In CXR No Finding, the full OOD set gives a strong positive correlation (Figure~\ref{fig:corr_summary}a), however, \texttt{OODSelect} retaining $>70\%$ of the data has a strong negative correlation (Figure~\ref{fig:corr_summary}). For Terra Incognita, the full OOD set has a strong positive correlation, but a 30\% slice from \texttt{OODSelect} has a notable negative correlation.

The extent of the existence of such subsets clearly varies across datasets and may not exist in others. For instance, for PACS, a small \texttt{OODSelect} size making up 8\% of the full dataset has a correlation of $-0.33$; at 60\,\% the correlation is already negligible, $0.01$, and becomes strongly positive as the size of \texttt{OODSelect} grows. The full dataset has a correlation of $0.81$.

Focusing specifically on the {most misclassified} examples, we find that the ID-OOD correlation is near $0$ and rarely invert it as the \texttt{OODSelect} examples do. This demonstrates that our selection does not conflate spuriousness with general difficulty (e.g., uniform label noise). Random selection consistently preserves a strong positive correlation similar to the full dataset, as expected. Thus \texttt{OODSelect} reveals systematic generalization failures due to spurious correlations, which may go undetected under standard evaluation across domains.

Clearly, to achieve a negative correlation from a positive correlation, we need (i) models that performed well on the full OOD set to perform relatively worse on the \texttt{OODSelect} set, and/or (ii) models that performed poorly on the full OOD set to perform relatively better on the \texttt{OODSelect} set. For instance, for VLCS, models with relatively low ID accuracy performed better on the \texttt{OODSelect} set than the full OOD set. In contrast, the models with relatively high ID accuracy performed better on the full OOD set. For all of our trends, the slope and intercept are determined by these relative performance changes and are dataset dependent. In some datasets, some models still perform near or below chance on the \texttt{OODSelect} (Terra Incognita) while in others, all models are above chance (WILDSCamelyon).

\begin{figure}[t]
    \centering
    \begin{subfigure}[t]{0.48\textwidth}
        \includegraphics[width=\linewidth]{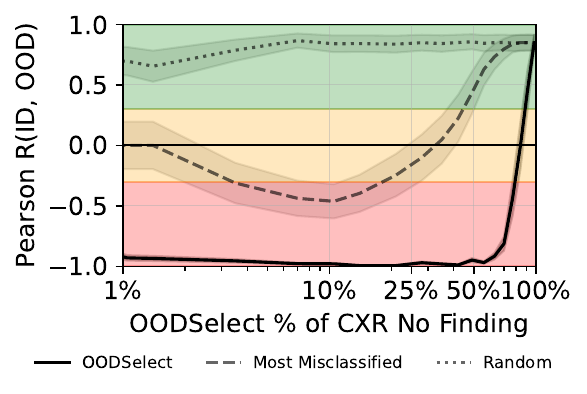} 
        \caption{CXR No Finding --- Pearson R.}
        \label{fig:cxr_pearson}
    \end{subfigure}
    \begin{subfigure}[t]{0.48\textwidth}
        \includegraphics[width=\linewidth]{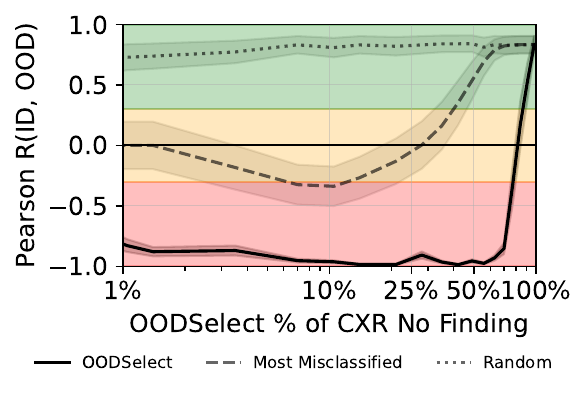} 
        \caption{CXR No Finding --- Spearman Rank.}
        \label{fig:cxr_spearman}
    \end{subfigure}
    \caption{The correlation directions are not driven by outliers --- Spearman rank is robust to outliers while Pearson R is not. Still, the trends are similar (full results in Figure~\ref{fig:corr_summary} and \ref{fig:spearman_corr_summary}).}
    \label{fig:cxr_outliers}
\end{figure}

\eparagraph{On the effect of outliers.} Some models may be outliers and skew the observed trends. Consequently, we evaluate Spearman rank correlation, which is more robust to outliers than Pearson $R$. We find that our conclusions remain unchanged (Figure~\ref{fig:cxr_outliers}). Spearman rank results are provided in Appendix~\ref{sec:empirical_details}. 

\begin{figure}[t]
    \centering
    \begin{subfigure}[t]{0.48\textwidth}
        \includegraphics[width=\linewidth]{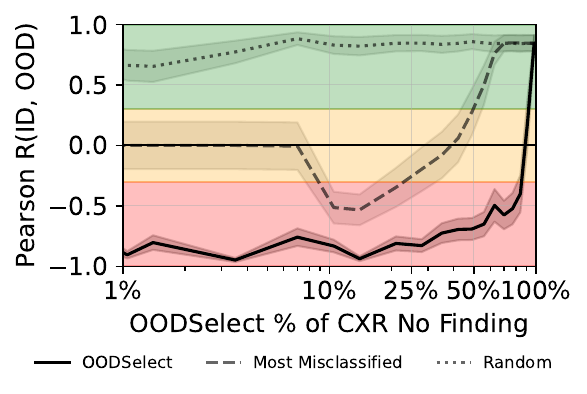} 
        \caption{Disjoint architecture families --- Pearson R.}
        \label{fig:arch_confound_pearson}
    \end{subfigure}
    \begin{subfigure}[t]{0.48\textwidth}
        \includegraphics[width=\linewidth]{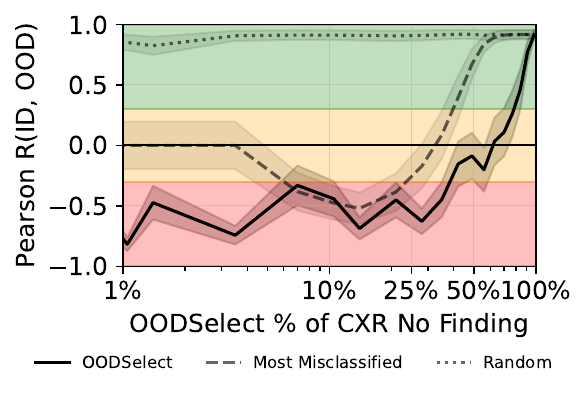} 
        \caption{Disjoint architecture families --- Spearman Rank.}
        \label{fig:arch_confound_spearman}
    \end{subfigure}
    \caption{Independent Architecture Families. Our findings hold even when train/test models are from disjoint architecture families, e.g., ResNets vs. ViTs.}
    \label{fig:arch_confound}
\end{figure}

\eparagraph{On potential architecture confounds.} While we randomly split models into disjoint sets for identifying OOD subsets and computing correlations to simulate i.i.d.\ sampling from a model population, architectural similarities (e.g., ResNet-50 vs.\ ResNet-152) could introduce confounding effects. To test this, we perform ablations where model families are disjoint—e.g., ResNets appear only in the training-validation set or only in the test set, but never both—and find that this restriction indeed changes the strength of the correlation, yet does not alter our conclusions. Figure~\ref{fig:arch_confound} gives an example for CXR No Finding, which has the strongest examples of AOTIL. However, given that architectures have different inductive biases, models may learn different spurious correlations or utilize them differently in decision-making. Sampling from an entirely disjoint population of architectures mitigates the observed strength of spurious correlations learned by model families.

\eparagraph{On Vision Language Models (VLM) Trends.} We investigate if the same trends hold with vision-language models' zero-shot performance~\citep{shi2024lca}. We generally find strongly positive correlations between the ID accuracy and the accuracy on \texttt{OODSelect} examples. The weakest correlations are: PACS (0.78), VLCS (0.62), TerraIncognita (0.84), WILDSCamelyon (1.00), and CXR (0.94). This should not be interpreted as evidence of VLMs' robustness to spurious correlations. From the point of view of the VLMs in this experiment, both the ID and \texttt{OODSelect} examples are OOD, since the VLMs were not explicitly trained on either set. Alternatively, since many of these datasets are publicly available, the VLMs may have been trained on the dataset sets, i.e., all of the examples are in distribution. 

\eparagraph{Selection via latent space distance.} As a baseline, we implement a selection method that greedily selects the farthest OOD examples from the ID examples in the CLIP embedding space~\citep{radford2021learning}. Across datasets, this approach often yields positive ID--OOD correlations (e.g., $R=0.52$ on PACS with $N=10$), and in some cases even stronger correlations than random selection. However, it consistently fails to capture the weak and negative correlations identified by OODSelect (e.g., $R=-0.92$ on VLCS with $N=10$). These results show that distance-based selection, while intuitive, overlooks the feature-label correlations that drive OOD errors, and thus cannot uncover the failure modes revealed by our method.

\begin{figure}
    \centering
    \includegraphics[width=0.75\linewidth]{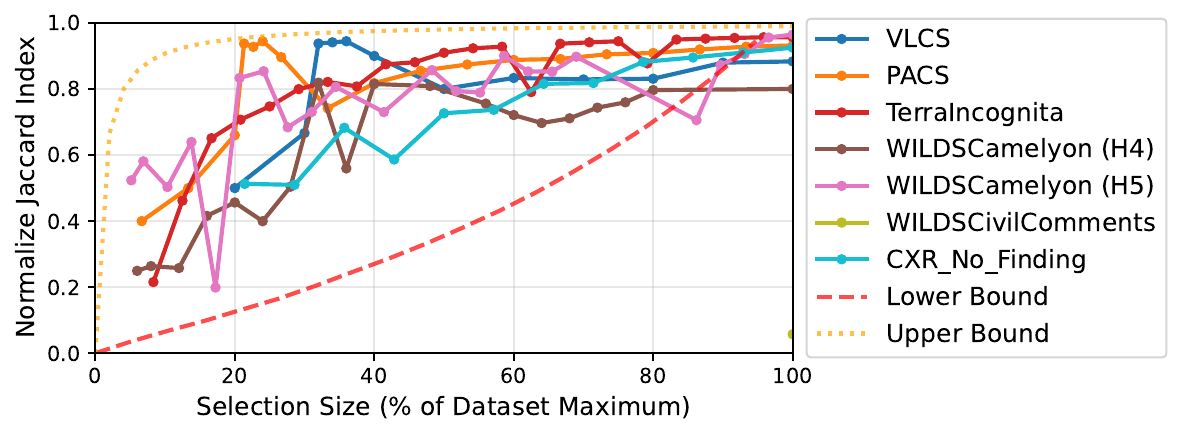}
    \caption{Consistency of selected subsets. Across all datasets and subset sizes, our normalized Jaccard index is greater than would be expected from arbitrary selection (lower bound).}
    \label{fig:jaccard_growth}
\end{figure}
\eparagraph{Selection Consistency and Coherence.} The identified subsets are also consistent and coherent. We select each $S$ subset independently and do not enforce that smaller subsets are subsets of larger subsets. Still, we find that such consistency holds. We measure consistency with the normalized \texttt{Jaccard Index}~\citep{jaccard1901etude} $\in [0, 1]$. For $\zb \subset [d]$ examples,
\begin{equation} \label{eq:jacc_idx}
   \Jcal(\zb_i, \zb_j) = \frac{\left|\zb_i \cap \zb_j \right|}{\left|\zb_i \cup \zb_j \right|}; \quad \quad \quad \bar{\Jcal}_\zb = \frac{1}{T}\sum_{k=1}^T \Jcal(\zb_k, \zb_{k+1}); \quad\quad\quad \tilde{\Jcal}_\zb = \frac{\bar{\Jcal}_{\zb} - \bar{\Jcal}_{\text{min}}}{\bar{\Jcal}_{\text{max}} - \bar{\Jcal}_{\text{min}}},
\end{equation}
for $T$ selection sizes, where $\tilde{\Jcal}_{\text{min}}$ is computed with random selections and $\tilde{\Jcal}_{\text{max}}$ with $\zb_i \subset \zb_j$ for all $i,j,$ with $i < j$, and the sizes of the sequence of $\zb_k$'s are preserved. This normalization is necessary since the subsets are of different sizes. Our selected subsets are indeed consistent (Figure~\ref{fig:jaccard_growth}). 

\eparagraph{CXR Semantic Coherence.}
\begin{table}[t]
    \caption{Model generated description of selected OOD set vs. ID set.}
    \label{tab:caption_diffs}
    \centering
    \begin{tabular}{lp{10.5cm}}
        \toprule
        \bfseries Dataset & \bfseries Model Generated Semantic Difference  \\
        \midrule
        PACS & extreme wide-angle shots; extreme close-ups; extreme weather conditions.\\
        VLCS & unusual object interactions with urban environments; x-ray or radiographic. \\ 
        TerraIncognita & frost; motion blur; extreme weather conditions; reflections or glare. \\ 
        
        \bottomrule
    \end{tabular}
\end{table}
The CXR dataset, predicting Finding/No Finding\footnote{``Finding'' indicates the presence of a condition from a predefined set and ``No Finding'' indicates otherwise.} in chest X-rays (CXR), is an example where we have demographic and clinical metadata that we can use to study the semantic coherence of our subsets. Figure~\ref{fig:cxr_selected_aotl} illustrates how average OOD performance can mask systematic failures in specific subsets. For instance, when selecting a subset with 5000 examples, Figure~\ref{fig:cxr_selected_aotl}, the ID/OOD accuracy correlation between ID and OOD on the selected subset is strongly negative, while it is strongly positive when we aggregate over the full OOD set.

We then analyze both demographic and clinical attributes. By comparing prevalence rates between the selected subset and the overall OOD pool, we find statistically significant shifts in several attributes, specifically \texttt{sex, race, Pleural Other, Support Devices, and Sex-Ethnicity}, determined via bootstrapping with 1000 resamples, Figure~\ref{fig:cxr_selected_aotl}. However, most datasets have no such metadata. Our normalized Jaccard Index supports consistency and coherence for such datasets.

\eparagraph{Potential for model-generated semantic coherence.} As a potential future research direction, we investigate the utility of large and vision language models to generate semantic concepts more likely to be true for our \texttt{OODSelect} set than the rest of the dataset~\citep{dunlap2024describing}. We apply the following process. {\em Step 1:} A VLM generates captions for all images; we use \texttt{Qwen2.5-32B-Instruct}~\citep{qwen2, qwen2.5}. {\em Step 2:} A large language model (LLM) then proposes candidate natural language descriptions that are more likely to apply to the selected OOD set than others; we use \texttt{AIMV2-large-patch14-224-lit}~\citep{fini2024multimodal}. {\em Step 3:} A vision-language model scores and ranks these descriptions based on their distinctiveness to OOD images, identifying interpretable attributes that differentiate the two distributions; we use \texttt{CLIP}~\citep{radford2021learning}.

Table~\ref{tab:caption_diffs} provides example descriptions for natural image datasets. We find that this strategy does not yield consistent and robust results, as the descriptions do not capture feature-label correlations, although some of our findings are promising. Additional details are provided in Appendix~\ref{sec:explanation}.

Importantly, many potentially spurious predictive features are incomprehensible to humans and may not be expressible in natural language~\citep{szegedy2013intriguing, goodfellow2014explaining, ilyas2019adversarial}. As a result, approaches that rely on vision-language models for explanation are likely still insufficient for identifying the subsets we uncover, for instance, through brute-force selection. They also cannot be expected to capture all differences in correlations between ID and OOD examples. This is true for many of the datasets in this work, such as CXR and WILDSCamelyon.

\newcommand{\meanerrpct}[2]{$#1\%_{\pm#2}$}

\eparagraph{Limitations.} Our analysis is computationally intensive, requiring training up to 4200 models per dataset and optimizing a selection objective of up to around 146000 elements. However, this computation is a one-time cost per dataset; we publicly release the resulting selections, covering many state-of-the-art domain generalization benchmarks. Additionally, semantic explanations of the \texttt{OODSelect} set are challenging for datasets such as WILDSCamelyon, whose features are images of tissue cell slides, without extensive metadata. Even when metadata is available, it may not fully represent the signals that capture spurious correlations. Notably, this is also an unstated challenge for the original datasets, where OOD sets are selected based on metadata such as hospital sites, but also contain no information explaining what spurious correlations exist or are expected. Furthermore, it is unclear if we can expect semantic explanations for all spurious correlations since many features models rely on are imperceptible to humans~\citep{szegedy2013intriguing, goodfellow2014explaining, ilyas2019adversarial}. For instance, AI systems can predict race from chest X-rays with features that are thus far imperceptible to humans~\citep{gichoya2022ai}.

\eparagraph{Broader Impact.} One alternative perspective of our results is that correlations that hold in aggregate are not spurious~\citep{wenzel2022assaying}. We propose that aggregate performance is a narrow view of the effect of spurious correlations. For instance, if spurious statistical associations reflecting historical or structural bias, such as occupation and gender, which can bias the outputs of recommendation systems~\citep{caliskan2017semantics, balagopalan2025s}, are pervasive in the real world. Then, benchmarks collected {\em naturally} from real-world distributions whose results are aggregated broadly may preserve such correlations across both training and test environments. As a result, models that rely on such spurious correlations may continue to ``perform well OOD,'' making the correlation appear benign in evaluation. However, this only creates the false impression that spurious correlations are not harmful OOD, even though they degrade performance on affected subsets of the data. Our work surfaces these subsets and advances more robust evaluations of OOD robustness.

\section{Conclusion}
Spurious correlations do not vanish in the real world; current benchmarks and performance metrics simply hide them through aggregation.  
By disaggregating OOD data, we revealed large, semantically meaningful subsets where spurious correlations harm performance. The consequential validity~\citep{messick1995validity, salaudeen2025measurement} of distribution shift robustness benchmarks, e.g., robustness to subpopulation shifts~\citep{yang2023change, sagawa2019distributionally}, requires identifying such subsets.

\eparagraph{Recommendations.} Future work in this area of research should (i) adopt our selection protocol as a robustness check for any new OOD benchmark, (ii) treat identified large \texttt{OODSelect} subsets as first-class evaluation targets, and (iii) design methods that improve \emph{both} average and subset robustness. A discussion on interpreting the results of subset performance can be found in~\cite{pfohl2025understanding}. We hope the released code and \texttt{OODSelect} subsets become a stepping stone toward benchmarks and models that confront the adverse effects of spurious correlations.

\FloatBarrier

\section*{Acknowledgments} MG acknowledges partial support by the National Science Foundation (NSF) 22-586 Faculty Early Career Development Award (\#2339381), a Gordon \& Betty Moore Foundation award, a Google Research Scholar award, and the AI2050 Program at Schmidt Sciences. We thank the anonymous reviewers of NeurIPS 2025 for their thoughtful feedback and recommendations. SB acknowledges partial support by the Schmidt Sciences AI2050 Program, NSF Awards No. 2330423 and 2441060, and NSERC Award No. 585136. Any opinions, findings and conclusions or recommendations expressed in this material are those of the author(s) and do not necessarily reflect the views of the NSF, NSERC, or Schmidt Sciences.
\bibliographystyle{unsrtnat}
\bibliography{main}

\newpage
\appendix
\section*{Appendix Table of Contents}
\startcontents[appendices]

\printcontents[appendices]{}{1}{\setcounter{tocdepth}{2}}
\FloatBarrier
\section{Empirical Analysis}\label{sec:empirical_details}
\paragraph{Vision Models.} \condensedVisionModelList 
\paragraph{Text Models.} \condensedTextModelList

\begin{figure}
        \centering
        \includegraphics[width=0.75\linewidth]{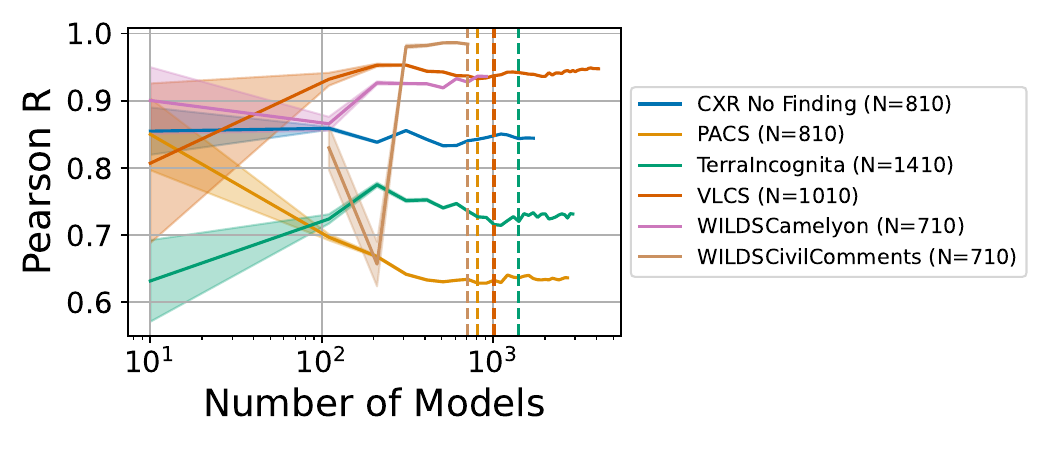}
    \caption{We train over 35 model architectures with varying hyperparameters, pretraining, and data augmentations, yielding an average of [N] trained models per dataset–OOD domain pair. We report the correlation between in-distribution (ID) and out-of-distribution (OOD) accuracy across all models, including standard errors at $\alpha = 0.05$. To ensure stability, we sample enough models such that adding more changes the correlation by less than 1\%; vertical dashed lines mark the approximate minimum sample size satisfying this criterion.}
    \label{fig:num_models}
\end{figure}

See Figure~\ref{fig:num_models} on quantity of models trained.

\paragraph{Dataset.} \textbf{PACS} involves object classification across artistic styles, with 7 classes (``dog'', ``elephant'', ``giraffe'', ``guitar'', ``horse'', ``house'', ``person'') and 4 domains: \textit{Photo}, \textit{Art Painting}, \textit{Cartoon}, and \textit{Sketch}. We consider a setting where \textit{Sketch} is the OOD domain. 
\textbf{VLCS} contains 5 object classes (``bird'', ``car'', ``chair'', ``dog'', ``person'') shared across 4 datasets: \textit{VOC2007}~\citep{everingham2010pascal}, \textit{LabelMe}~\citep{russell2008labelme}, \textit{Caltech101}~\citep{fei2004learning}, and \textit{SUN09}~\citep{choi2010exploiting}. Each domain reflects a different dataset source with distinct collection biases. We consider a setting where \textit{LabelMe} is the OOD domain. 
\textbf{Terra Incognita} focuses on wildlife recognition from camera trap images, with 10 classes (``bird'', ``bobcat'', ``cat'', ``coyote'', ``dog'', ``opossum'', ``raccoon'', ``rabbit'', ``skunk'', ``squirrel'') across 4 geographically distinct domains: \textit{L38}, \textit{L43}, \textit{L46}, and \textit{L100}. The \textit{L46} location is the OOD domain~\citep{gulrajani2020search}.

We also study three WILDS benchmarks that capture distinct real-world distribution shifts. We consider WILDS-Camelyon~\citep{bandi2018detection} and WILDS-CivilComments~\citep{borkan2019nuanced}. These benchmarks encompass medical imaging, satellite vision, and natural language, providing a diverse evaluation suite for real-world generalization under domain shift. \textbf{WILDS-Camelyon} is a histopathology image classification task for detecting cancerous regions in lymph node slides, with domain shifts arising from variations across different hospitals. \textbf{WILDS-CivilComments}~\citep{borkan2019nuanced} classifies online comments as toxic or non-toxic across demographic subgroups (\textit{male}, \textit{female}, \textit{LGBTQ}, \textit{Christian}, \textit{Muslim}, \textit{other religions}, \textit{Black}, \textit{White}), with OOD domains reflecting shifts in identity distributions.

\textbf{CXR (``Finding'' vs.\ ``No Finding'').}%
\footnote{Throughout, we refer to this task as \textit{CXR} for brevity.} This binary classification task predicts whether a chest X-ray shows any abnormal radiological finding.   The in-distribution (ID) domains comprise four widely-used datasets—\textit{CheXpert} (v1.0-small) \citep{irvin2019chexpert}, \textit{ChestXray8} \citep{wang2017chestx}, \textit{PadChest} \citep{bustos2020padchest}, and \textit{VinDr-CXR} \citep{nguyen2022vindr}.   These sources differ in scanner hardware, patient demographics, annotation guidelines, and prevalence of pathologies. We designate \textit{MIMIC-CXR} \citep{johnson2019mimic}—a large, single-institution dataset collected under a distinct clinical workflow—as the out-of-distribution (OOD) domain. This setting captures clinically meaningful shifts (e.g., hospital protocols, imaging devices, disease prevalence) and offers a stringent test of real-world generalization under domain shift.

\eparagraph{Train/Val/Test Split.} To evaluate generalization, we randomly partition the same set of models into train, validation, and test splits (60/20/20). We optimize our selection objective on the training split and identify the best-performing \texttt{OODSelect} configuration using the held-out validation split. Final results are reported on the test split. Importantly, although the selection objective is tuned on one subset of models, the ID and OOD accuracy correlations continue to hold on the held-out test models, demonstrating that the property generalizes across held-out model subsets.

\eparagraph{Soundness of the relaxation and optimization.}
Notably, our objective is non‑convex and non‑submodular (Proposition~\ref{prop:nonsubmodularity}) yet Lipschitz‑continuous (Lemma~\ref{lem:pearson_lipschitz}).  Consequently, while global optimality is intractable, the Lipschitz property ensures that gradient‑based methods with a suitably large exact‑penalty parameter admit meaningful descent guarantees; in practice we employ stochastic gradient descent with multiple random restarts, which consistently converges to high‑quality feasible solutions.  Formal optimization guarantees and proofs are deferred to Appendix~\ref{sec:theory}.

Adding the squared regularization term in~\eqref{eq:main_objective} is an {\em exact‑penalty} reformulation of the original constrained problem.  
Classical results~\citep{bertsekas1997nonlinear, nocedal1999numerical}) state that there exists a finite weight $\lambda^\star>0$ such that, for every $\lambda\!\ge\!\lambda^\star$, (i) every global minimiser of the penalised objective satisfies the budget constraint, and (ii) the optimal value coincides with that of the constrained problem.

Moreover, any first‑order stationary point that already meets the constraint is {\em unchanged} by the penalty term, so the relaxation does not create spurious local optima within the feasible region. Hence, gradient‑based search on~\eqref{eq:main_objective} is sufficient: we do not need to solve the penalized problem to global optimality, and any locally optimal feasible solution we find is also locally optimal for the original constrained objective.

We use a cosine annealing schedule to adjust the learning rate and  $\lambda$~\citep{loshchilov2016sgdr}. Additional details are available in Appendix~\ref{sec:empirical_details}.

\eparagraph{Necessary Number of Models.} Prior work on accuracy-on-the-line typically trained only tens of models per dataset~\citep{miller2021accuracy}. In contrast, we train thousands of models per dataset, spanning architectures from AlexNet to Vision Transformers and incorporating diverse training strategies. To determine how many models are necessary for stable correlation estimates, we incrementally sample models until the Pearson correlation between ID and OOD accuracies changes by less than 1\%. Figure~\ref{fig:num_models} shows where this stability threshold is reached for each dataset. Across all datasets, our experiments far exceed this threshold.

\eparagraph{On the Size of \texttt{OODSelect}.} While a detailed analysis of thresholding \texttt{OODSelect} is beyond our current scope, we generally recommend choosing the largest \texttt{OODSelect} size such that the Pearson correlation is not weak, that is $R\le -0.3$, following convention~\citep{cohen2013statistical}. For some datasets, this threshold may yield very small or noisy selections (e.g., PACS), in which case the selected set may not be informative.

\subsection{Compute}
\begin{table}[h]
\caption{Compute time to reproduce experiments (GPU Hours) --- per experiment unit on NVIDIA RTX A6000 GPUs.}
\label{tab:compute}
\centering
\begin{tabular}{lccccccc}
\toprule
Dataset & mean & median & std. dev. & min & max & total\\
\midrule
CXR & 4 & 3 & 3 & <1 & 18 & 286 \\
PACS & 1 & 1 & 1 & <1 & 7 & 109 \\
TerraIncognita & 2 & 2 & 2 & <1 & 10 & 292 \\
VLCS & 2 & 2 & 1 & <1 & 7 & 183 \\
WILDSCamelyon & 4 & 2 & 4 & 1 & 18 & 350 \\
WILDSCivilComments & 10 & 9 & 4 & 3 & 18 & 106 \\
\bottomrule
\end{tabular}
\end{table}

\subsection{Spearman Rank Results}
\begin{figure}[h!]
    \centering
    \includegraphics[width=\textwidth]{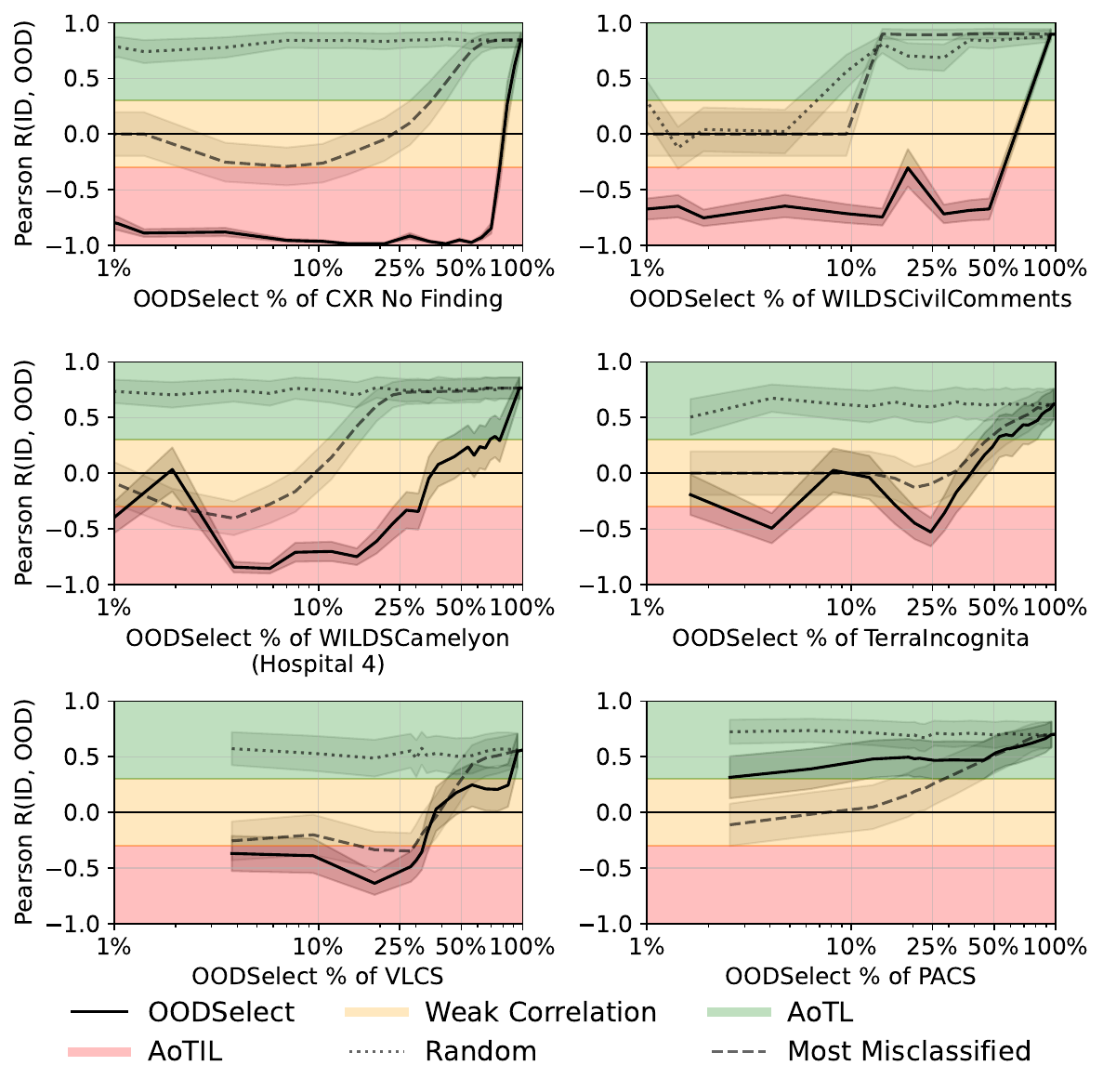}
    \caption{\textbf{Comparing AoTL and AoTIL.} Spearman Correlation between ID and OOD accuracy as a function of the number of selected OOD samples.}
    \label{fig:spearman_corr_summary}
\end{figure}

\begin{landscape}
\clearpage
\subsection{Table of Results}
{\small
\begin{longtable}{l c c c c c c c c}
\caption{ID vs. selected OOD accuracy correlations (Pearson $R$ and Spearman $\rho$) with standard errors over 100 resamplings. For ``Random'', we randomly select $N$ subsets from candidate OOD examples; for ``Hard'', we select the $N$ most misclassified examples.}
\label{tab:corrs}\\

\toprule
\textbf{Dataset} & \textbf{OOD} & \textbf{N} &
\multicolumn{3}{c}{\textbf{Pearson $R$}} &
\multicolumn{3}{c}{\textbf{Spearman $\rho$}} \\
\cmidrule(lr{0.5em}){4-6}\cmidrule(lr{0.5em}){7-9}
 &  &  &
\textbf{Ours} & \textbf{Random} & \textbf{Hard} &
\textbf{Ours} & \textbf{Random} & \textbf{Hard} \\
\midrule
\endfirsthead

\toprule
\textbf{Dataset} & \textbf{OOD} & \textbf{N} &
\multicolumn{3}{c}{\textbf{Pearson $R$}} &
\multicolumn{3}{c}{\textbf{Spearman $\rho$}} \\
\cmidrule(lr{0.5em}){4-6}\cmidrule(lr{0.5em}){7-9}
 &  &  &
\textbf{Ours} & \textbf{Random} & \textbf{Hard} &
\textbf{Ours} & \textbf{Random} & \textbf{Hard} \\
\midrule
\endhead

\midrule
\multicolumn{9}{r}{Continued on next page} \\
\midrule
\endfoot

\bottomrule
\endlastfoot

CXR No Finding & MIMIC-CXR & 10 & -0.56 (0.12) & 0.19 (0.20) & 0.00 (0.20) & -0.46 (0.14) & 0.07 (0.20) & 0.00 (0.20) \\
CXR No Finding & MIMIC-CXR & 20 & -0.23 (0.18) & 0.20 (0.20) & 0.00 (0.20) & -0.33 (0.16) & 0.14 (0.20) & 0.00 (0.20) \\
CXR No Finding & MIMIC-CXR & 50 & -0.59 (0.12) & 0.43 (0.18) & 0.00 (0.20) & -0.58 (0.12) & 0.46 (0.17) & 0.00 (0.20) \\
CXR No Finding & MIMIC-CXR & 100 & -0.69 (0.09) & 0.59 (0.14) & 0.00 (0.20) & -0.60 (0.11) & 0.52 (0.16) & 0.00 (0.20) \\
CXR No Finding & MIMIC-CXR & 250 & -0.83 (0.05) & 0.70 (0.11) & 0.00 (0.20) & -0.72 (0.08) & 0.57 (0.15) & 0.00 (0.20) \\
CXR No Finding & MIMIC-CXR & 500 & -0.84 (0.05) & 0.71 (0.11) & 0.00 (0.20) & -0.70 (0.09) & 0.72 (0.11) & 0.00 (0.20) \\
CXR No Finding & MIMIC-CXR & 750 & -0.93 (0.02) & 0.69 (0.12) & 0.00 (0.20) & -0.81 (0.06) & 0.75 (0.10) & 0.00 (0.20) \\
CXR No Finding & MIMIC-CXR & 1000 & -0.93 (0.02) & 0.66 (0.13) & 0.00 (0.20) & -0.87 (0.04) & 0.72 (0.11) & 0.00 (0.20) \\
CXR No Finding & MIMIC-CXR & 2500 & -0.95 (0.02) & 0.78 (0.09) & -0.30 (0.17) & -0.86 (0.04) & 0.78 (0.09) & -0.19 (0.18) \\
CXR No Finding & MIMIC-CXR & 5000 & -0.98 (0.01) & 0.86 (0.06) & -0.40 (0.15) & -0.95 (0.02) & 0.84 (0.07) & -0.29 (0.17) \\
CXR No Finding & MIMIC-CXR & 7500 & -0.98 (0.01) & 0.84 (0.07) & -0.44 (0.15) & -0.96 (0.01) & 0.83 (0.07) & -0.30 (0.17) \\
CXR No Finding & MIMIC-CXR & 10000 & -0.99 (0.00) & 0.84 (0.07) & -0.37 (0.16) & -0.99 (0.00) & 0.84 (0.07) & -0.25 (0.18) \\
CXR No Finding & MIMIC-CXR & 15000 & -0.99 (0.00) & 0.84 (0.07) & -0.22 (0.18) & -0.98 (0.00) & 0.83 (0.07) & -0.13 (0.19) \\
CXR No Finding & MIMIC-CXR & 20000 & -0.97 (0.01) & 0.85 (0.07) & -0.07 (0.19) & -0.91 (0.03) & 0.84 (0.07) & 0.02 (0.20) \\
CXR No Finding & MIMIC-CXR & 25000 & -0.98 (0.01) & 0.84 (0.07) & 0.08 (0.20) & -0.96 (0.01) & 0.84 (0.07) & 0.18 (0.20) \\
CXR No Finding & MIMIC-CXR & 30000 & -0.99 (0.00) & 0.85 (0.07) & 0.26 (0.19) & -0.99 (0.00) & 0.85 (0.07) & 0.37 (0.18) \\
CXR No Finding & MIMIC-CXR & 35000 & -0.95 (0.02) & 0.85 (0.06) & 0.45 (0.17) & -0.95 (0.01) & 0.85 (0.07) & 0.56 (0.15) \\
CXR No Finding & MIMIC-CXR & 40000 & -0.97 (0.01) & 0.84 (0.07) & 0.64 (0.13) & -0.98 (0.01) & 0.83 (0.07) & 0.71 (0.11) \\
CXR No Finding & MIMIC-CXR & 45000 & -0.91 (0.03) & 0.85 (0.07) & 0.74 (0.10) & -0.93 (0.02) & 0.84 (0.07) & 0.80 (0.08) \\
CXR No Finding & MIMIC-CXR & 50000 & -0.83 (0.05) & 0.85 (0.07) & 0.80 (0.08) & -0.87 (0.04) & 0.84 (0.07) & 0.83 (0.07) \\
CXR No Finding & MIMIC-CXR & 55000 & -0.46 (0.14) & 0.84 (0.07) & 0.84 (0.07) & -0.32 (0.17) & 0.84 (0.07) & 0.84 (0.07) \\
CXR No Finding & MIMIC-CXR & 60000 & -0.02 (0.20) & 0.84 (0.07) & 0.85 (0.07) & 0.18 (0.20) & 0.84 (0.07) & 0.84 (0.07) \\
CXR No Finding & MIMIC-CXR & 65000 & 0.46 (0.17) & 0.85 (0.07) & 0.85 (0.07) & 0.56 (0.15) & 0.84 (0.07) & 0.84 (0.07) \\
CXR No Finding & MIMIC-CXR & 70000 & 0.85 (0.07) & 0.85 (0.07) & 0.85 (0.07) & 0.84 (0.07) & 0.84 (0.07) & 0.84 (0.07) \\
CXR No Finding & MIMIC-CXR & 71433 & 0.85 & 0.85 & 0.85 & 0.84 & 0.84 & 0.84 \\
\hline
WILDSCivilComments & 4 & 10 & -0.33 (0.16) & 0.39 (0.18) & 0.00 (0.20) & -0.50 (0.13) & 0.16 (0.20) & 0.00 (0.20) \\
WILDSCivilComments & 4 & 20 & -0.89 (0.03) & 0.85 (0.06) & 0.00 (0.20) & -0.26 (0.17) & 0.13 (0.20) & 0.00 (0.20) \\
WILDSCivilComments & 4 & 50 & -0.89 (0.03) & 0.39 (0.18) & 0.00 (0.20) & -0.55 (0.12) & 0.23 (0.19) & 0.00 (0.20) \\
WILDSCivilComments & 4 & 100 & -0.56 (0.12) & 0.74 (0.10) & 0.00 (0.20) & -0.53 (0.13) & 0.19 (0.20) & 0.00 (0.20) \\
WILDSCivilComments & 4 & 250 & -0.79 (0.06) & 0.94 (0.03) & 0.00 (0.20) & -0.71 (0.09) & 0.33 (0.19) & 0.00 (0.20) \\
WILDSCivilComments & 4 & 500 & -0.70 (0.09) & 0.93 (0.03) & 0.00 (0.20) & -0.75 (0.08) & 0.46 (0.17) & 0.00 (0.20) \\
WILDSCivilComments & 4 & 750 & -0.98 (0.01) & 0.80 (0.08) & 0.00 (0.20) & -0.65 (0.10) & 0.02 (0.20) & 0.00 (0.20) \\
WILDSCivilComments & 4 & 1000 & -0.93 (0.02) & 0.82 (0.08) & 0.00 (0.20) & -0.82 (0.06) & 0.08 (0.20) & 0.00 (0.20) \\
WILDSCivilComments & 4 & 2500 & -0.78 (0.07) & 0.93 (0.03) & 0.00 (0.20) & -0.74 (0.08) & -0.01 (0.20) & 0.00 (0.20) \\
WILDSCivilComments & 4 & 5000 & -0.98 (0.01) & 0.95 (0.02) & 0.00 (0.20) & -0.76 (0.07) & 0.47 (0.17) & 0.00 (0.20) \\
WILDSCivilComments & 4 & 7500 & -0.98 (0.01) & 0.97 (0.01) & 0.95 (0.03) & -0.80 (0.06) & 0.79 (0.09) & 0.88 (0.05) \\
WILDSCivilComments & 4 & 10000 & -0.88 (0.04) & 0.97 (0.01) & 0.99 (0.01) & -0.41 (0.15) & 0.70 (0.12) & 0.90 (0.05) \\
WILDSCivilComments & 4 & 15000 & -0.97 (0.01) & 0.97 (0.02) & 0.98 (0.01) & -0.78 (0.07) & 0.70 (0.12) & 0.89 (0.05) \\
WILDSCivilComments & 4 & 20000 & -0.97 (0.01) & 0.98 (0.01) & 0.98 (0.01) & -0.76 (0.07) & 0.83 (0.07) & 0.90 (0.05) \\
WILDSCivilComments & 4 & 25000 & -0.96 (0.01) & 0.98 (0.01) & 0.98 (0.01) & -0.73 (0.08) & 0.84 (0.07) & 0.91 (0.04) \\
WILDSCivilComments & 4 & 50000 & 0.98 (0.01) & 0.98 (0.01) & 0.98 (0.01) & 0.90 (0.05) & 0.90 (0.05) & 0.90 (0.05) \\
WILDSCivilComments & 4 & 52823 & 0.98 & 0.98 & 0.98 & 0.90 & 0.90 & 0.90 \\
\hline
WILDSCamelyon & Hospital 4 & 10 & -0.94 (0.02) & 0.90 (0.05) & 0.00 (0.20) & -0.23 (0.18) & 0.33 (0.19) & 0.00 (0.20) \\
WILDSCamelyon & Hospital 4 & 20 & -0.91 (0.03) & 0.74 (0.11) & 0.00 (0.20) & -0.17 (0.18) & 0.25 (0.19) & 0.00 (0.20) \\
WILDSCamelyon & Hospital 4 & 50 & -0.93 (0.02) & 0.49 (0.16) & 0.00 (0.20) & -0.32 (0.17) & 0.50 (0.16) & 0.00 (0.20) \\
WILDSCamelyon & Hospital 4 & 100 & -0.92 (0.03) & 0.94 (0.03) & 0.00 (0.20) & -0.34 (0.16) & 0.52 (0.16) & 0.00 (0.20) \\
WILDSCamelyon & Hospital 4 & 250 & -0.88 (0.04) & 0.96 (0.02) & 0.00 (0.20) & -0.42 (0.15) & 0.37 (0.18) & 0.00 (0.20) \\
WILDSCamelyon & Hospital 4 & 500 & -0.96 (0.01) & 0.97 (0.02) & 0.00 (0.20) & -0.53 (0.13) & 0.49 (0.17) & 0.00 (0.20) \\
WILDSCamelyon & Hospital 4 & 750 & -0.96 (0.01) & 0.98 (0.01) & 0.00 (0.20) & -0.61 (0.11) & 0.75 (0.10) & 0.00 (0.20) \\
WILDSCamelyon & Hospital 4 & 1000 & -0.91 (0.03) & 0.99 (0.01) & 0.00 (0.20) & -0.51 (0.13) & 0.74 (0.10) & 0.00 (0.20) \\
WILDSCamelyon & Hospital 4 & 2500 & 0.07 (0.20) & 0.98 (0.01) & 0.08 (0.20) & 0.00 (0.20) & 0.71 (0.11) & -0.51 (0.13) \\
WILDSCamelyon & Hospital 4 & 5000 & -0.99 (0.00) & 0.99 (0.01) & -0.32 (0.17) & -0.93 (0.02) & 0.77 (0.09) & -0.31 (0.17) \\
WILDSCamelyon & Hospital 4 & 7500 & -0.98 (0.01) & 0.99 (0.01) & -0.23 (0.18) & -0.95 (0.02) & 0.78 (0.09) & -0.24 (0.18) \\
WILDSCamelyon & Hospital 4 & 10000 & -0.98 (0.01) & 0.98 (0.01) & -0.10 (0.19) & -0.88 (0.04) & 0.77 (0.10) & -0.09 (0.19) \\
WILDSCamelyon & Hospital 4 & 15000 & -0.97 (0.01) & 0.99 (0.01) & 0.23 (0.19) & -0.82 (0.06) & 0.78 (0.09) & 0.23 (0.19) \\
WILDSCamelyon & Hospital 4 & 20000 & -0.95 (0.02) & 0.99 (0.01) & 0.52 (0.16) & -0.80 (0.06) & 0.79 (0.09) & 0.48 (0.17) \\
WILDSCamelyon & Hospital 4 & 25000 & -0.88 (0.04) & 0.99 (0.01) & 0.70 (0.12) & -0.49 (0.14) & 0.80 (0.08) & 0.64 (0.13) \\
WILDSCamelyon & Hospital 4 & 30000 & -0.90 (0.03) & 0.99 (0.01) & 0.83 (0.07) & -0.50 (0.13) & 0.80 (0.09) & 0.76 (0.10) \\
WILDSCamelyon & Hospital 4 & 35000 & -0.54 (0.13) & 0.99 (0.01) & 0.92 (0.04) & -0.28 (0.17) & 0.80 (0.08) & 0.79 (0.09) \\
WILDSCamelyon & Hospital 4 & 40000 & -0.08 (0.19) & 0.99 (0.01) & 0.98 (0.01) & -0.37 (0.16) & 0.78 (0.09) & 0.79 (0.09) \\
WILDSCamelyon & Hospital 4 & 45000 & 0.08 (0.20) & 0.99 (0.01) & 0.98 (0.01) & -0.02 (0.20) & 0.78 (0.09) & 0.79 (0.09) \\
WILDSCamelyon & Hospital 4 & 50000 & 0.27 (0.19) & 0.99 (0.01) & 0.98 (0.01) & 0.12 (0.20) & 0.81 (0.08) & 0.79 (0.09) \\
WILDSCamelyon & Hospital 4 & 60000 & 0.30 (0.19) & 0.99 (0.01) & 0.98 (0.01) & 0.17 (0.20) & 0.78 (0.09) & 0.80 (0.08) \\
WILDSCamelyon & Hospital 4 & 70000 & 0.36 (0.18) & 0.99 (0.01) & 0.98 (0.01) & 0.24 (0.19) & 0.80 (0.09) & 0.80 (0.08) \\
WILDSCamelyon & Hospital 4 & 75000 & 0.40 (0.18) & 0.99 (0.01) & 0.98 (0.01) & 0.18 (0.20) & 0.79 (0.09) & 0.80 (0.08) \\
WILDSCamelyon & Hospital 4 & 80000 & 0.41 (0.18) & 0.99 (0.01) & 0.98 (0.01) & 0.24 (0.19) & 0.80 (0.08) & 0.80 (0.08) \\
WILDSCamelyon & Hospital 4 & 85000 & 0.45 (0.17) & 0.99 (0.01) & 0.98 (0.01) & 0.23 (0.19) & 0.81 (0.08) & 0.80 (0.08) \\
WILDSCamelyon & Hospital 4 & 90000 & 0.47 (0.17) & 0.99 (0.01) & 0.99 (0.01) & 0.28 (0.19) & 0.80 (0.08) & 0.81 (0.08) \\
WILDSCamelyon & Hospital 4 & 95000 & 0.50 (0.16) & 0.99 (0.01) & 0.99 (0.01) & 0.29 (0.19) & 0.81 (0.08) & 0.81 (0.08) \\
WILDSCamelyon & Hospital 4 & 100000 & 0.91 (0.04) & 0.99 (0.01) & 0.99 (0.01) & 0.26 (0.19) & 0.80 (0.08) & 0.81 (0.08) \\
WILDSCamelyon & Hospital 4 & 125000 & 0.99 (0.01) & 0.99 (0.01) & 0.99 (0.01) & 0.81 (0.08) & 0.81 (0.08) & 0.81 (0.08) \\
WILDSCamelyon & Hospital 4 & 129838 & 0.99 & 0.99 & 0.99 & 0.81 & 0.81 & 0.81 \\
WILDSCamelyon & Hospital 5 & 10 & -0.63 (0.10) & 0.25 (0.19) & 0.00 (0.20) & -0.03 (0.19) & 0.63 (0.14) & 0.00 (0.20) \\
WILDSCamelyon & Hospital 5 & 20 & -0.83 (0.05) & 0.43 (0.18) & 0.00 (0.20) & 0.13 (0.20) & 0.48 (0.17) & 0.00 (0.20) \\
WILDSCamelyon & Hospital 5 & 50 & -0.74 (0.08) & 0.73 (0.11) & 0.00 (0.20) & 0.16 (0.20) & 0.39 (0.18) & 0.00 (0.20) \\
WILDSCamelyon & Hospital 5 & 100 & -0.81 (0.06) & 0.71 (0.11) & -0.18 (0.18) & 0.21 (0.20) & 0.36 (0.18) & -0.26 (0.17) \\
WILDSCamelyon & Hospital 5 & 250 & -0.91 (0.03) & 0.75 (0.10) & -0.18 (0.18) & -0.53 (0.13) & 0.48 (0.17) & -0.46 (0.14) \\
WILDSCamelyon & Hospital 5 & 500 & -0.87 (0.04) & 0.82 (0.08) & -0.51 (0.13) & 0.33 (0.19) & 0.36 (0.18) & -0.63 (0.11) \\
WILDSCamelyon & Hospital 5 & 750 & -0.81 (0.06) & 0.80 (0.08) & -0.45 (0.14) & 0.36 (0.18) & 0.47 (0.17) & -0.64 (0.10) \\
WILDSCamelyon & Hospital 5 & 1000 & -0.83 (0.05) & 0.84 (0.07) & -0.38 (0.16) & 0.35 (0.18) & 0.57 (0.15) & -0.59 (0.11) \\
WILDSCamelyon & Hospital 5 & 2500 & -0.61 (0.11) & 0.80 (0.08) & -0.48 (0.14) & 0.26 (0.19) & 0.46 (0.17) & -0.74 (0.08) \\
WILDSCamelyon & Hospital 5 & 5000 & -0.43 (0.15) & 0.80 (0.08) & -0.44 (0.15) & 0.40 (0.18) & 0.49 (0.16) & -0.63 (0.11) \\
WILDSCamelyon & Hospital 5 & 7500 & -0.44 (0.15) & 0.81 (0.08) & -0.38 (0.16) & -0.11 (0.19) & 0.47 (0.17) & -0.51 (0.13) \\
WILDSCamelyon & Hospital 5 & 10000 & -0.68 (0.09) & 0.81 (0.08) & -0.32 (0.17) & -0.39 (0.15) & 0.48 (0.17) & -0.41 (0.15) \\
WILDSCamelyon & Hospital 5 & 15000 & -0.58 (0.12) & 0.80 (0.08) & -0.15 (0.19) & -0.71 (0.09) & 0.47 (0.17) & -0.22 (0.18) \\
WILDSCamelyon & Hospital 5 & 20000 & -0.84 (0.05) & 0.81 (0.08) & 0.01 (0.20) & 0.02 (0.20) & 0.48 (0.17) & 0.01 (0.20) \\
WILDSCamelyon & Hospital 5 & 25000 & -0.81 (0.06) & 0.80 (0.08) & 0.14 (0.20) & 0.09 (0.20) & 0.47 (0.17) & 0.17 (0.20) \\
WILDSCamelyon & Hospital 5 & 30000 & -0.77 (0.07) & 0.81 (0.08) & 0.25 (0.19) & 0.15 (0.20) & 0.48 (0.17) & 0.28 (0.19) \\
WILDSCamelyon & Hospital 5 & 35000 & -0.68 (0.09) & 0.81 (0.08) & 0.32 (0.19) & 0.23 (0.19) & 0.48 (0.17) & 0.35 (0.18) \\
WILDSCamelyon & Hospital 5 & 40000 & -0.71 (0.08) & 0.80 (0.08) & 0.38 (0.18) & 0.23 (0.19) & 0.47 (0.17) & 0.40 (0.18) \\
WILDSCamelyon & Hospital 5 & 45000 & -0.20 (0.18) & 0.80 (0.08) & 0.43 (0.17) & 0.18 (0.20) & 0.47 (0.17) & 0.41 (0.18) \\
WILDSCamelyon & Hospital 5 & 50000 & -0.66 (0.10) & 0.80 (0.08) & 0.48 (0.17) & 0.27 (0.19) & 0.47 (0.17) & 0.42 (0.18) \\
WILDSCamelyon & Hospital 5 & 60000 & -0.62 (0.11) & 0.80 (0.08) & 0.54 (0.16) & 0.28 (0.19) & 0.47 (0.17) & 0.43 (0.17) \\
WILDSCamelyon & Hospital 5 & 70000 & 0.13 (0.20) & 0.81 (0.08) & 0.59 (0.15) & 0.28 (0.19) & 0.48 (0.17) & 0.46 (0.17) \\
WILDSCamelyon & Hospital 5 & 75000 & 0.29 (0.19) & 0.80 (0.08) & 0.67 (0.12) & 0.33 (0.19) & 0.48 (0.17) & 0.47 (0.17) \\
WILDSCamelyon & Hospital 5 & 80000 & 0.31 (0.19) & 0.81 (0.08) & 0.73 (0.11) & 0.38 (0.18) & 0.48 (0.17) & 0.47 (0.17) \\
WILDSCamelyon & Hospital 5 & 85000 & 0.40 (0.18) & 0.80 (0.08) & 0.77 (0.09) & 0.35 (0.18) & 0.48 (0.17) & 0.47 (0.17) \\
WILDSCamelyon & Hospital 5 & 90000 & 0.37 (0.18) & 0.80 (0.08) & 0.79 (0.09) & 0.40 (0.18) & 0.48 (0.17) & 0.47 (0.17) \\
WILDSCamelyon & Hospital 5 & 95000 & 0.40 (0.18) & 0.80 (0.08) & 0.81 (0.08) & 0.43 (0.18) & 0.48 (0.17) & 0.47 (0.17) \\
WILDSCamelyon & Hospital 5 & 100000 & 0.56 (0.15) & 0.80 (0.08) & 0.82 (0.08) & 0.41 (0.18) & 0.48 (0.17) & 0.47 (0.17) \\
WILDSCamelyon & Hospital 5 & 125000 & 0.66 (0.13) & 0.81 (0.08) & 0.82 (0.08) & 0.45 (0.17) & 0.48 (0.17) & 0.48 (0.17) \\
WILDSCamelyon & Hospital 5 & 130000 & 0.70 (0.12) & 0.80 (0.08) & 0.81 (0.08) & 0.42 (0.18) & 0.48 (0.17) & 0.48 (0.17) \\
WILDSCamelyon & Hospital 5 & 135000 & 0.72 (0.11) & 0.81 (0.08) & 0.81 (0.08) & 0.43 (0.17) & 0.48 (0.17) & 0.48 (0.17) \\
WILDSCamelyon & Hospital 5 & 140000 & 0.76 (0.10) & 0.81 (0.08) & 0.81 (0.08) & 0.44 (0.17) & 0.48 (0.17) & 0.48 (0.17) \\
WILDSCamelyon & Hospital 5 & 145000 & 0.81 (0.08) & 0.80 (0.08) & 0.80 (0.08) & 0.48 (0.17) & 0.48 (0.17) & 0.48 (0.17) \\
WILDSCamelyon & Hospital 5 & 146722 & 0.80 & 0.80 & 0.80 & 0.48 & 0.48 & 0.48 \\
\hline
TerraIncognita & L46 & 10 & -0.86 (0.04) & 0.45 (0.17) & 0.00 (0.20) & -0.36 (0.16) & 0.45 (0.17) & 0.00 (0.20) \\
TerraIncognita & L46 & 20 & -0.90 (0.03) & 0.92 (0.04) & 0.00 (0.20) & -0.36 (0.16) & 0.49 (0.17) & 0.00 (0.20) \\
TerraIncognita & L46 & 50 & 0.40 (0.18) & 0.92 (0.04) & 0.00 (0.20) & -0.18 (0.18) & 0.44 (0.17) & 0.00 (0.20) \\
TerraIncognita & L46 & 100 & -0.58 (0.12) & 0.87 (0.06) & 0.00 (0.20) & -0.33 (0.16) & 0.46 (0.17) & 0.00 (0.20) \\
TerraIncognita & L46 & 250 & -0.91 (0.03) & 0.90 (0.04) & 0.00 (0.20) & -0.46 (0.14) & 0.66 (0.13) & 0.00 (0.20) \\
TerraIncognita & L46 & 500 & -0.77 (0.07) & 0.87 (0.06) & 0.03 (0.20) & -0.05 (0.19) & 0.58 (0.15) & 0.02 (0.20) \\
TerraIncognita & L46 & 750 & -0.74 (0.08) & 0.89 (0.05) & 0.12 (0.20) & -0.01 (0.20) & 0.59 (0.15) & -0.03 (0.20) \\
TerraIncognita & L46 & 1000 & -0.77 (0.07) & 0.89 (0.05) & -0.24 (0.18) & -0.22 (0.18) & 0.61 (0.14) & -0.12 (0.19) \\
TerraIncognita & L46 & 1250 & -0.59 (0.11) & 0.90 (0.04) & -0.24 (0.18) & -0.43 (0.15) & 0.60 (0.14) & -0.15 (0.19) \\
TerraIncognita & L46 & 1500 & -0.40 (0.15) & 0.90 (0.04) & -0.22 (0.18) & -0.54 (0.13) & 0.58 (0.15) & -0.10 (0.19) \\
TerraIncognita & L46 & 1750 & -0.26 (0.17) & 0.87 (0.06) & -0.19 (0.18) & -0.38 (0.16) & 0.60 (0.14) & -0.06 (0.19) \\
TerraIncognita & L46 & 2000 & -0.12 (0.19) & 0.89 (0.05) & -0.12 (0.19) & -0.23 (0.18) & 0.62 (0.14) & 0.02 (0.20) \\
TerraIncognita & L46 & 2250 & -0.08 (0.19) & 0.89 (0.05) & -0.05 (0.19) & -0.12 (0.19) & 0.59 (0.14) & 0.08 (0.20) \\
TerraIncognita & L46 & 2500 & -0.02 (0.20) & 0.89 (0.05) & 0.04 (0.20) & -0.00 (0.20) & 0.61 (0.14) & 0.17 (0.20) \\
TerraIncognita & L46 & 2750 & 0.04 (0.20) & 0.89 (0.05) & 0.11 (0.20) & 0.10 (0.20) & 0.59 (0.14) & 0.25 (0.19) \\
TerraIncognita & L46 & 3000 & 0.09 (0.20) & 0.90 (0.05) & 0.18 (0.20) & 0.17 (0.20) & 0.60 (0.14) & 0.32 (0.19) \\
TerraIncognita & L46 & 3250 & 0.15 (0.20) & 0.89 (0.05) & 0.25 (0.19) & 0.27 (0.19) & 0.59 (0.14) & 0.36 (0.18) \\
TerraIncognita & L46 & 3500 & 0.25 (0.19) & 0.90 (0.05) & 0.30 (0.19) & 0.25 (0.19) & 0.62 (0.14) & 0.39 (0.18) \\
TerraIncognita & L46 & 3750 & 0.29 (0.19) & 0.89 (0.05) & 0.37 (0.18) & 0.24 (0.19) & 0.60 (0.14) & 0.43 (0.18) \\
TerraIncognita & L46 & 4000 & 0.34 (0.19) & 0.90 (0.05) & 0.43 (0.17) & 0.31 (0.19) & 0.61 (0.14) & 0.45 (0.17) \\
TerraIncognita & L46 & 4250 & 0.45 (0.17) & 0.90 (0.04) & 0.51 (0.16) & 0.36 (0.18) & 0.61 (0.14) & 0.47 (0.17) \\
TerraIncognita & L46 & 4500 & 0.58 (0.15) & 0.89 (0.05) & 0.59 (0.14) & 0.36 (0.18) & 0.61 (0.14) & 0.49 (0.16) \\
TerraIncognita & L46 & 4750 & 0.64 (0.13) & 0.89 (0.05) & 0.66 (0.13) & 0.38 (0.18) & 0.59 (0.14) & 0.52 (0.16) \\
TerraIncognita & L46 & 5000 & 0.70 (0.12) & 0.89 (0.05) & 0.72 (0.11) & 0.41 (0.18) & 0.61 (0.14) & 0.54 (0.15) \\
TerraIncognita & L46 & 5250 & 0.75 (0.10) & 0.89 (0.05) & 0.77 (0.09) & 0.48 (0.17) & 0.61 (0.14) & 0.54 (0.16) \\
TerraIncognita & L46 & 5500 & 0.80 (0.08) & 0.89 (0.05) & 0.81 (0.08) & 0.52 (0.16) & 0.60 (0.14) & 0.57 (0.15) \\
TerraIncognita & L46 & 5750 & 0.84 (0.07) & 0.89 (0.05) & 0.85 (0.07) & 0.55 (0.15) & 0.60 (0.14) & 0.60 (0.14) \\
TerraIncognita & L46 & 6000 & 0.89 (0.05) & 0.89 (0.05) & 0.88 (0.05) & 0.61 (0.14) & 0.60 (0.14) & 0.60 (0.14) \\
TerraIncognita & L46 & 6122 & 0.89 & 0.89 & 0.89 & 0.60 & 0.60 & 0.60 \\
\hline
VLCS & LabelMe & 10 & -0.91 (0.03) & 0.80 (0.08) & 0.00 (0.20) & -0.28 (0.17) & 0.16 (0.20) & 0.00 (0.20) \\
VLCS & LabelMe & 20 & -0.90 (0.03) & 0.90 (0.04) & 0.00 (0.20) & -0.30 (0.17) & 0.27 (0.19) & 0.00 (0.20) \\
VLCS & LabelMe & 50 & -0.07 (0.19) & 0.93 (0.03) & -0.13 (0.19) & -0.29 (0.17) & 0.43 (0.18) & -0.15 (0.19) \\
VLCS & LabelMe & 100 & -0.82 (0.05) & 0.96 (0.02) & -0.22 (0.18) & -0.39 (0.15) & 0.57 (0.15) & -0.21 (0.18) \\
VLCS & LabelMe & 250 & -0.27 (0.17) & 0.94 (0.03) & -0.25 (0.18) & -0.37 (0.16) & 0.57 (0.15) & -0.28 (0.17) \\
VLCS & LabelMe & 500 & -0.40 (0.15) & 0.94 (0.03) & -0.21 (0.18) & -0.62 (0.11) & 0.52 (0.16) & -0.29 (0.17) \\
VLCS & LabelMe & 750 & -0.40 (0.15) & 0.95 (0.03) & -0.25 (0.18) & -0.50 (0.13) & 0.59 (0.15) & -0.33 (0.16) \\
VLCS & LabelMe & 800 & -0.33 (0.16) & 0.94 (0.03) & -0.23 (0.18) & -0.43 (0.15) & 0.51 (0.16) & -0.26 (0.17) \\
VLCS & LabelMe & 850 & -0.28 (0.17) & 0.95 (0.03) & -0.18 (0.18) & -0.33 (0.16) & 0.61 (0.14) & -0.18 (0.18) \\
VLCS & LabelMe & 900 & -0.17 (0.18) & 0.95 (0.03) & -0.10 (0.19) & -0.17 (0.18) & 0.56 (0.15) & -0.10 (0.19) \\
VLCS & LabelMe & 1000 & 0.09 (0.20) & 0.95 (0.03) & 0.03 (0.20) & 0.08 (0.20) & 0.59 (0.15) & 0.03 (0.20) \\
VLCS & LabelMe & 1250 & 0.33 (0.19) & 0.94 (0.03) & 0.47 (0.17) & 0.23 (0.19) & 0.55 (0.15) & 0.28 (0.19) \\
VLCS & LabelMe & 1500 & 0.65 (0.13) & 0.95 (0.03) & 0.73 (0.11) & 0.30 (0.19) & 0.55 (0.15) & 0.49 (0.16) \\
VLCS & LabelMe & 1750 & 0.81 (0.08) & 0.95 (0.03) & 0.87 (0.06) & 0.26 (0.19) & 0.60 (0.14) & 0.56 (0.15) \\
VLCS & LabelMe & 2000 & 0.87 (0.06) & 0.95 (0.03) & 0.92 (0.04) & 0.26 (0.19) & 0.62 (0.14) & 0.58 (0.15) \\
VLCS & LabelMe & 2250 & 0.91 (0.04) & 0.95 (0.02) & 0.94 (0.03) & 0.31 (0.19) & 0.62 (0.14) & 0.59 (0.14) \\
VLCS & LabelMe & 2500 & 0.95 (0.03) & 0.95 (0.02) & 0.94 (0.03) & 0.60 (0.14) & 0.61 (0.14) & 0.60 (0.14) \\
VLCS & LabelMe & 2656 & 0.95 & 0.95 & 0.95 & 0.61 & 0.61 & 0.61 \\
\hline
PACS & Sketch & 10 & -0.48 (0.14) & 0.37 (0.18) & 0.17 (0.20) & -0.05 (0.19) & 0.39 (0.18) & -0.06 (0.19) \\
PACS & Sketch & 20 & -0.33 (0.16) & 0.71 (0.11) & 0.17 (0.20) & -0.41 (0.15) & 0.34 (0.19) & 0.00 (0.20) \\
PACS & Sketch & 50 & -0.47 (0.14) & 0.84 (0.07) & 0.00 (0.20) & 0.23 (0.19) & 0.47 (0.17) & -0.12 (0.19) \\
PACS & Sketch & 100 & -0.33 (0.16) & 0.73 (0.11) & 0.11 (0.20) & 0.29 (0.19) & 0.70 (0.12) & -0.08 (0.19) \\
PACS & Sketch & 250 & -0.30 (0.17) & 0.79 (0.09) & 0.19 (0.20) & 0.35 (0.19) & 0.70 (0.12) & -0.04 (0.19) \\
PACS & Sketch & 500 & -0.22 (0.18) & 0.83 (0.07) & 0.28 (0.19) & 0.41 (0.18) & 0.69 (0.12) & 0.07 (0.20) \\
PACS & Sketch & 750 & 0.01 (0.20) & 0.82 (0.08) & 0.33 (0.19) & 0.43 (0.17) & 0.65 (0.13) & 0.18 (0.20) \\
PACS & Sketch & 800 & 0.05 (0.20) & 0.81 (0.08) & 0.33 (0.19) & 0.42 (0.18) & 0.66 (0.13) & 0.19 (0.20) \\
PACS & Sketch & 850 & 0.06 (0.20) & 0.80 (0.08) & 0.33 (0.19) & 0.42 (0.18) & 0.64 (0.13) & 0.22 (0.20) \\
PACS & Sketch & 900 & 0.08 (0.20) & 0.83 (0.07) & 0.34 (0.19) & 0.42 (0.18) & 0.65 (0.13) & 0.24 (0.19) \\
PACS & Sketch & 1000 & 0.10 (0.20) & 0.80 (0.08) & 0.35 (0.19) & 0.40 (0.18) & 0.67 (0.12) & 0.27 (0.19) \\
PACS & Sketch & 1250 & 0.16 (0.20) & 0.81 (0.08) & 0.38 (0.18) & 0.41 (0.18) & 0.67 (0.12) & 0.33 (0.19) \\
PACS & Sketch & 1500 & 0.21 (0.20) & 0.82 (0.08) & 0.41 (0.18) & 0.42 (0.18) & 0.68 (0.12) & 0.39 (0.18) \\
PACS & Sketch & 1750 & 0.24 (0.19) & 0.82 (0.08) & 0.46 (0.17) & 0.42 (0.18) & 0.67 (0.12) & 0.43 (0.17) \\
PACS & Sketch & 2000 & 0.29 (0.19) & 0.82 (0.08) & 0.49 (0.16) & 0.48 (0.17) & 0.66 (0.13) & 0.48 (0.17) \\
PACS & Sketch & 2250 & 0.48 (0.17) & 0.81 (0.08) & 0.54 (0.16) & 0.52 (0.16) & 0.67 (0.12) & 0.51 (0.16) \\
PACS & Sketch & 2500 & 0.56 (0.15) & 0.81 (0.08) & 0.58 (0.15) & 0.54 (0.16) & 0.67 (0.12) & 0.56 (0.15) \\
PACS & Sketch & 2750 & 0.62 (0.14) & 0.81 (0.08) & 0.63 (0.13) & 0.56 (0.15) & 0.67 (0.13) & 0.61 (0.14) \\
PACS & Sketch & 3000 & 0.67 (0.12) & 0.81 (0.08) & 0.68 (0.12) & 0.58 (0.15) & 0.67 (0.12) & 0.64 (0.13) \\
PACS & Sketch & 3250 & 0.71 (0.11) & 0.81 (0.08) & 0.72 (0.11) & 0.61 (0.14) & 0.67 (0.13) & 0.66 (0.13) \\
PACS & Sketch & 3500 & 0.75 (0.10) & 0.81 (0.08) & 0.76 (0.10) & 0.63 (0.14) & 0.66 (0.13) & 0.66 (0.13) \\
PACS & Sketch & 3750 & 0.81 (0.08) & 0.81 (0.08) & 0.79 (0.09) & 0.67 (0.13) & 0.66 (0.13) & 0.67 (0.13) \\
PACS & Sketch & 3929 & 0.81 & 0.81 & 0.81 & 0.67 & 0.67 & 0.67 \\

\bottomrule
\end{longtable}
}
\end{landscape}

\FloatBarrier
\section{Theoretical Analysis} \label{sec:theory}
\begin{figure}
    \centering
    \begin{subfigure}[t]{0.45\textwidth}
        \includegraphics[width=\linewidth]{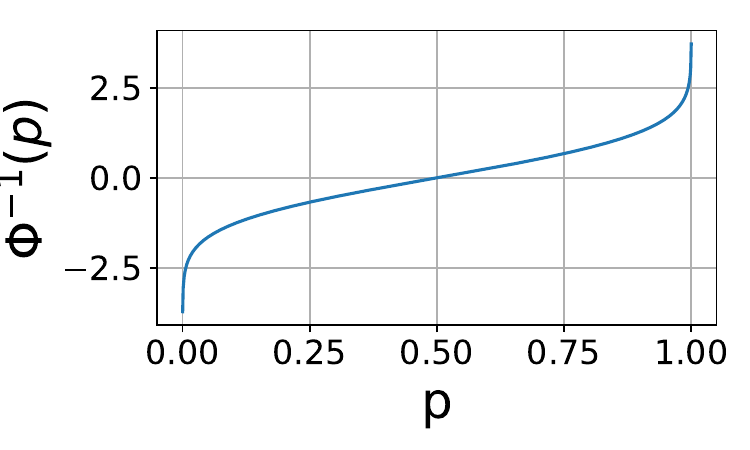}
        \caption{Inverse CDF (Probit) of the Standard Normal}
    \end{subfigure}
    \hfill
    \begin{subfigure}[t]{0.45\textwidth}
        \includegraphics[width=\linewidth]{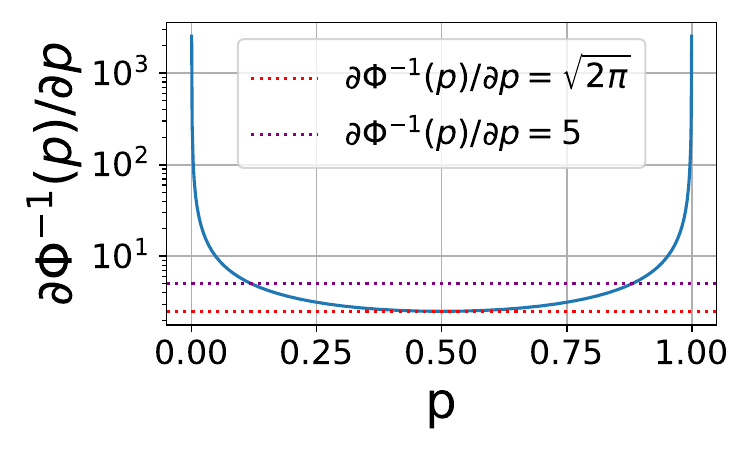}
        \caption{Derivative of the Inverse Normal CDF $= 1 / \phi(\Phi^{-1}(p))$ where $\phi$ and $\Phi$ are PDF and CDF, respectively.}
    \end{subfigure}
    \caption{The probit transform is indeed Lipschitz.}
    \label{fig:logit_domain}
\end{figure}

\subsection{Lemma~\ref{lem:bnd_model}: Bounded effect of New Models on Pearson R}
\begin{lemma}[Bounded Effect of a New Model on Pearson $R$]\label{lem:bnd_model}
Let $(\zb_i,\wb_i)_{i=1}^{N}\subseteq[\alpha,1-\alpha]^2$ with $\alpha\in(0,1)$.
Define
\[\xb_i=\Phi^{-1}(\zb_i), \qquad \yb_i=\Phi^{-1}(\wb_i),
\]
and let
\[
\rho_N=\text{corr}\bigl(\xb_{1:N},\yb_{1:N}\bigr)
\]
be the sample Pearson correlation of the first $N$ transformed pairs.
Add one more pair $(\zb_{N+1},\wb_{N+1})$ with $\zb_{N+1}\in[\alpha,1-\alpha]$
and $\wb_{N+1}=\beta\,\zb_{N+1}$, and denote the updated correlation by
$\rho_{N+1}$.  Then
\[
\bigl|\rho_{N+1}-\rho_N\bigr| \,\le\, \frac{\kappa(1+|\beta|)\,M_\alpha^{2}}{N}, \qquad M_\alpha=\max\!\bigl\{\lvert\Phi^{-1}(\alpha)\rvert,\lvert\Phi^{-1}(1-\alpha)\rvert\bigr\},
\]
where the constant $\kappa>0$ depends only on $\alpha$
(via the Lipschitz constant of $\Phi^{-1}$ on $[\alpha,1-\alpha]$).
\begin{proof}
Let $\xb_i = \Phi^{-1}(\zb_i)$ and $\yb_i = \Phi^{-1}(\wb_i)$ for $i = 1, \dots, N$. Denote $\bar{\xb}_N := \frac{1}{N} \sum_{i=1}^N \xb_i$ and similarly $\bar{\yb}_N$. Let $\rho_N$ be the Pearson correlation between $(\xb_1, \dots, \xb_N)$ and $(\yb_1, \dots, \yb_N)$:
\[
\rho_N = \frac{\sum_{i=1}^N (\xb_i - \bar{\xb}_N)(\yb_i - \bar{\yb}_N)}{\sqrt{\sum_{i=1}^N (\xb_i - \bar{\xb}_N)^2 \sum_{i=1}^N (\yb_i - \bar{\yb}_N)^2}}.
\]
Let $(\zb_{N+1}, \wb_{N+1})$ be a new pair with $\wb_{N+1} = \beta \zb_{N+1}$ and both in $[\alpha, 1 - \alpha]$. Define $\xb_{N+1} = \Phi^{-1}(\zb_{N+1})$ and $\yb_{N+1} = \Phi^{-1}(\wb_{N+1})$. Since $\Phi^{-1}$ is $L_\alpha$-Lipschitz on $[\alpha, 1 - \alpha]$, we have $|\xb_{N+1}| \le M_\alpha$ and $|\yb_{N+1}| \le L_\alpha |\beta| + M_\alpha \le (1 + |\beta|) M_\alpha$. Thus, each $|\xb_i|, |\yb_i| \le (1 + |\beta|) M_\alpha$.

Let $\rho_{N+1}$ denote the Pearson correlation after adding $(\xb_{N+1}, \yb_{N+1})$. A first-order perturbation of the sample Pearson correlation (cf. derivative bounds on correlation statistics) gives:
\[
|\rho_{N+1} - \rho_N| \le \frac{C}{N} \cdot \max_{i} \{|\xb_i|, |\yb_i|\}^2,
\]
where $C$ is a constant depending on the Lipschitz constant $L_\alpha$ and lower bound on variance (which is lower bounded by $(\alpha(1-\alpha)/L_\alpha^2)$ due to the probit transform). Therefore,
\[
|\rho_{N+1} - \rho_N| \le \frac{\kappa(1 + |\beta|) M_\alpha^2}{N},
\]
where $\kappa > 0$ depends only on $\alpha$.
\end{proof}
\end{lemma}

\subsection{Lemma~\ref{lem:bnd_example}: Bounded effect of New Examples on Pearson R}
\begin{lemma}[Bounded Effect of a New Example on Pearson $R$]\label{lem:bnd_example}
Fix $\alpha\in(0,1)$ and assume the per-model accuracies satisfy
$\zb_i^{(S)},\,\wb_i\in[\alpha,1-\alpha]$ for $i=1,\dots,d$.
Define 
\[\xb_i=\Phi^{-1}(\zb_i), \qquad \yb_i=\Phi^{-1}(\wb_i),
\]
and let
\[
\rho_S=\text{corr}\!\bigl(\tilde{\xb}^{(S)},\tilde{\yb}\bigr)
\]
be their sample Pearson correlation. Now add an additional selected example. Note that each average changes by at most $|\zb_i^{(S+1)}-\zb_i^{(S)}|\le 1/(S+1).$ Write $\rho_{S+1}$ for the resulting correlation between $\xb^{(S+1)}$ and $\yb$. Then
\[
\bigl|\rho_{S+1}-\rho_S\bigr| \,\le\, \frac{\kappa\,M_\alpha^{2}}{S+1}, \qquad M_\alpha=\max\!\bigl\{\lvert\Phi^{-1}(\alpha)\rvert, \lvert\Phi^{-1}(1-\alpha)\rvert\bigr\},
\]
where the constant $\kappa>0$ depends only on $\alpha$
(via the Lipschitz constant of $\Phi^{-1}$ on $[\alpha,1-\alpha]$).
\begin{proof}
Let $\xb_i = \Phi^{-1}(\zb_i^{(S)})$, $\yb_i = \Phi^{-1}(\wb_i)$, and let $\rho_S = \text{corr}(\tilde{\xb}, \tilde{\yb})$. Adding a new example changes each $\zb_i^{(S)}$ by at most $1/(S+1)$, so $|\xb_i^{(S+1)} - \xb_i^{(S)}| \le L_\alpha / (S+1)$. Denote $\delta_i = \xb_i^{(S+1)} - \xb_i^{(S)}$ and $\bar{\delta} = \frac{1}{d} \sum_{i=1}^d \delta_i$. Then
\[
|\tilde{\xb}_i^{(S+1)} - \tilde{\xb}_i^{(S)}| = |\delta_i - \bar{\delta}| \le 2 L_\alpha / (S+1).
\]
Using $\|\tilde{\yb}\|_\infty \le M_\alpha$,
\[
\left| \tilde{\xb}^{(S+1)\top} \tilde{\yb} - \tilde{\xb}^{(S)\top} \tilde{\yb} \right| \le \sum_{i=1}^d |\tilde{\xb}_i^{(S+1)} - \tilde{\xb}_i^{(S)}||\tilde{\yb}_i| \le \frac{2d L_\alpha M_\alpha}{S+1}.
\]
Also, $\|\tilde{\xb}\|, \|\tilde{\yb}\| \ge \sqrt{d v_\alpha}$ for $v_\alpha := \alpha(1-\alpha) L_\alpha^{-2}$. So
\[
|\rho_{S+1} - \rho_S| \le \frac{2 d L_\alpha M_\alpha}{(S+1) d v_\alpha} = \frac{2 L_\alpha M_\alpha}{(S+1) v_\alpha} = \frac{2 L_\alpha^3 M_\alpha}{\alpha(1-\alpha)(S+1)}.
\]
Setting $\kappa := \frac{2 L_\alpha^3}{\alpha(1-\alpha)}$ gives the result.
\end{proof}

\end{lemma}

\subsection{Lemma~\ref{lem:pearson_lipschitz}: Lipschitz Continuity of Selection Objective}
\begin{lemma}[Lipschitz Continuity of Pearson Correlation w.r.t.\ $\sbb$]
\label{lem:pearson_lipschitz}
Let $\Xb \in \mathbb{R}^{N\times d}$ be the binary accuracy matrix and $\yb\in\mathbb{R}^N$ the held-out training accuracy vector. Define the test-set accuracy for a given selection vector $\sbb\in[0,1]^d$ (with fixed total mass $\sum_{j=1}^d \sbb_j = S>0$) as 
\[
\hat{\xb}(\sbb)=\frac{\Xb\,\sbb}{S},
\]
and let $f$ be a Lipschitz-continuous probit transformation with Lipschitz constant $L_f$. Denote
\[
\tilde{\xb}(\sbb)=f\bigl(\hat{\xb}(\sbb)\bigr), \quad \tilde{\yb}=f(\yb),
\]
and the centered versions by
\[
\bar{\xb}(\sbb)=\tilde{\xb}(\sbb)-\frac{1}{N}\one^\top\tilde{\xb}(\sbb), \quad \bar{\yb}=\tilde{\yb}-\frac{1}{N}\one^\top\tilde{\yb}.
\]
The Pearson correlation between $\tilde{\xb}(\sbb)$ and $\tilde{\yb}$ is defined as
\[
\text{corr}\bigl(\tilde{\xb}(\sbb),\tilde{\yb}\bigr)
=\frac{\frac{1}{N}\bar{\xb}(\sbb)^\top\bar{\yb}}{\sqrt{\frac{1}{N^2}\|\bar{\xb}(\sbb)\|^2\,\|\bar{\yb}\|^2}}.
\]
Assume that there exists $\epsilon>0$ such that $\|\bar{\xb}(\sbb)\|\ge \epsilon$ for all admissible $\sbb$. Then, for any two selection vectors $\sbb, \sbb'\in[0,1]^d$ with $\sum_{j=1}^d \sbb_j = \sum_{j=1}^d \sbb'_j = S$, there exists a constant $L>0$ (depending on $L_f$, $\Xb$, $S$, and $\epsilon$) such that
\[
\Bigl|\text{corr}\bigl(\tilde{\xb}(\sbb),\tilde{\yb}\bigr)-
\text{corr}\bigl(\tilde{\xb}(\sbb'),\tilde{\yb}\bigr)\Bigr|
\le L\,\|\sbb-\sbb'\|.
\]
\end{lemma}

\begin{proof}
Since the test-set accuracy is given by
\[
\hat{\xb}(\sbb)=\frac{\Xb\,\sbb}{S},
\]
linearity implies that
\[
\|\hat{\xb}(\sbb)-\hat{\xb}(\sbb')\|
\le \frac{\|\Xb\|}{S}\,\|\sbb-\sbb'\|,
\]
where $\|\Xb\|$ denotes an appropriate operator norm.

Using the Lipschitz continuity of $f$ (with constant $L_f$), we have for each coordinate,
\[
\Bigl|f\bigl(\hat{x}_i(\sbb)\bigr)-f\bigl(\hat{x}_i(\sbb')\bigr)\Bigr|
\le L_f\,\Bigl|\hat{x}_i(\sbb)-\hat{x}_i(\sbb')\Bigr|,
\]
so in vector form,
\[
\|\tilde{\xb}(\sbb)-\tilde{\xb}(\sbb')\|
\le L_f\,\|\hat{\xb}(\sbb)-\hat{\xb}(\sbb')\|
\le \frac{L_f\,\|\Xb\|}{S}\,\|\sbb-\sbb'\|.
\]
Since centering is a linear operation, it follows that
\[
\|\bar{\xb}(\sbb)-\bar{\xb}(\sbb')\|
\le \|\tilde{\xb}(\sbb)-\tilde{\xb}(\sbb')\|
\le \frac{L_f\,\|\Xb\|}{S}\,\|\sbb-\sbb'\|.
\]

Now, note that the Pearson correlation is computed as
\[
\text{corr}\bigl(\tilde{\xb}(\sbb),\tilde{\yb}\bigr)
=\frac{\bar{\xb}(\sbb)^\top\bar{\yb}}{\|\bar{\xb}(\sbb)\|\,\|\bar{\yb}\|}.
\]
Since $\bar{\yb}$ is independent of $\sbb$ and by the assumption that $\|\bar{\xb}(\sbb)\|\ge \epsilon>0$, standard arguments (via the mean value theorem and the differentiability of the quotient function on a compact domain) imply that there exists a constant $C>0$ such that
\[
\Bigl|\text{corr}\bigl(\tilde{\xb}(\sbb),\tilde{\yb}\bigr)-
\text{corr}\bigl(\tilde{\xb}(\sbb'),\tilde{\yb}\bigr)\Bigr|
\le C\,\|\bar{\xb}(\sbb)-\bar{\xb}(\sbb')\|.
\]
Thus, combining the bounds yields
\[
\Bigl|\text{corr}\bigl(\tilde{\xb}(\sbb),\tilde{\yb}\bigr)-
\text{corr}\bigl(\tilde{\xb}(\sbb'),\tilde{\yb}\bigr)\Bigr|
\le C\,\frac{L_f\,\|\Xb\|}{S}\,\|\sbb-\sbb'\|.
\]
Setting $L = C\frac{L_f\,\|\Xb\|}{S}$ completes the proof.
\end{proof}

\subsection{Proof of Proposition~\ref{prop:nonsubmodularity}: Non-Submodularity}
\begin{proposition}[Non-Submodularity; no diminishing returns (Informal)]\label{prop:nonsubmodularity}
    Let $\sbb \in \{0,1\}^d$ be a selection vector over $d$ candidate OOD examples, and write
    $\corr({\accidb}, {\acc}_{\OOD}^{\sbb})$ for the Pearson correlation in Eq.~\eqref{eq:corr}.
    Define $\sbb_i \preceq \sbb_j$ to mean $(\sbb_i)_t \le (\sbb_j)_t$ for all $t\in\{1,\ldots,d\}$.
    For $k\in\{1,\ldots,d\}$ with $(\sbb)_k=0$, let $\sbb^{+k}$ denote the vector obtained by setting the $k$th coordinate of $\sbb$ to $1$ (leaving all others unchanged).
    Then, in general, there exist $\sbb_i,\sbb_j \in \{0,1\}^d$ with $\sbb_i \preceq \sbb_j$ and $(\sbb_i)_k=(\sbb_j)_k=0$ such that
    \[
        \corr\!\big({\accidb},\, {\acc}_{\OOD}^{\sbb_i^{+k}}\big)
        \;-\;
        \corr\!\big({\accidb},\, {\acc}_{\OOD}^{\sbb_i}\big)
        \;<\;
        \corr\!\big({\accidb},\, {\acc}_{\OOD}^{\sbb_j^{+k}}\big)
        \;-\;
        \corr\!\big({\accidb},\, {\acc}_{\OOD}^{\sbb_j}\big).
    \]
\end{proposition}

Let $\Mcal(\sbb) = \text{corr}\left(\xb_\sbb, y\right)$. The exists $\sbb_i \subseteq \sbb_j \subseteq \{1, \ldots, d\}$ and $j \ne \sbb_j$ such that $$\Mcal(\sbb_i \cup \{j\}) - \Mcal(\sbb_i) < \Mcal(\sbb_j \cup \{j\}) - \Mcal(\sbb_j).$$

\begin{proof}
Let $y\in\RR^n$ satisfy $\|y-\bar y\mathbf 1\|_2>0$.  
Choose three candidate columns
\[
\xb_1 \;=\; y,\qquad
\xb_2 \;\text{independent of }y\text{ with } \text{corr}(\xb_2,y)=0,\qquad
\xb_3 \;=\; y .
\]
Set the index sets
\[
\sbb_i=\{1\},\quad
\sbb_j=\{1,2\},\quad
j=3 .
\]

Compute the four correlations:
\[
\Mcal(\sbb_i)=\text{corr}(\xb_1,y)=1,\qquad
\Mcal(\sbb_i\cup\{3\})=\text{corr}\!\Bigl(\tfrac{\xb_1+\xb_3}{2},\,y\Bigr)=1,
\]
\[
\Mcal(\sbb_j)=\text{corr}\!\Bigl(\tfrac{\xb_1+\xb_2}{2},\,y\Bigr)=\tfrac12,
\qquad
\Mcal(\sbb_j\cup\{3\})=\text{corr}\!\Bigl(\tfrac{\xb_1+\xb_2+\xb_3}{3},\,y\Bigr)=\tfrac23 .
\]

\noindent
Hence
\[
\underbrace{\Mcal(\sbb_i\cup\{3\})-\Mcal(\sbb_i)}_{=\,0}
\;<\;
\underbrace{\Mcal(\sbb_j\cup\{3\})-\Mcal(\sbb_j)}_{=\,\frac16},
\]
so $\Mcal$ violates the diminishing-returns (submodularity) condition.
\end{proof}

\FloatBarrier
\section{ID/OOD Explanation}\label{sec:explanation}
\begin{algorithm}[h]
\caption{Descriptive Differences Between ID and OOD Subsets~\citep{dunlap2024describing}}
\label{algo:caption_differences}
\KwIn{ID image set $\Dcal_{\ID}$, OOD image set $\Dcal_{\OOD}$}
\KwOut{Ranked list of difference descriptions highlighting OOD-specific attributes}

\textbf{Step 1: Generate Captions}\;
Use BLIP-2~\cite{li2023blip} to generate captions for all images in $\Dcal_{\ID}$ and $\Dcal_{\OOD}$\;

\textbf{Step 2: Generate Candidate Difference Descriptions}\;
Use Mixtral~\cite{jiang2024mixtral} to propose a set of natural language descriptions more likely to appear in $\Dcal_{\OOD}$ than in $\Dcal_{\ID}$, based on the generated captions\;

\textbf{Step 3: Score and Rank Differences}\;
\ForEach{difference description $d$}{
    Compute average CLIP~\cite{radford2021learning} similarity between $d$ and images in $\Dcal_{\ID}$: $\text{sim}_{\ID}(d)$\;
    Compute average CLIP similarity between $d$ and images in $\Dcal_{\OOD}$: $\text{sim}_{\OOD}(d)$\;
    Compute difference score: $\Delta(d) = \text{sim}_{\OOD}(d) - \text{sim}_{\ID}(d)$\;
}
Rank all descriptions by $\Delta(d)$ in descending order\;

\Return{Top-$k$ difference descriptions ranked by distinctiveness to $\Dcal_{\OOD}$}\;
\end{algorithm}


We follow Algorithm~\ref{algo:caption_differences} to generate semantic difference between selected OOD samples and ID samples. We select 200 samples from the ID and selected OOD sets, respectively. As a motivation for future work, we enumerate our observations from this cursory study.

We evaluated multiple vision-language and language models to generate and summarize conceptual differences between In-Distribution (ID) and Out-of-Distribution (OOD) image groups.

\paragraph{Similarity Scoring Models.} We tested CLIP, SigLIP, and AimV2 (CVPR 2025) to score image-caption similarity. AimV2 produced the most meaningful rankings based on manual inspection.

\paragraph{Prompting Strategy.} Initial difference captions often described \textit{groups} of images (e.g., ``images with different views''), which misaligned with how CLIP-like models are trained (single image-caption pairs). To address this:
\begin{itemize}
    \item We introduced a detailed prompt discouraging group-level descriptions. Larger models responded well, while smaller models were sensitive and inconsistent.
    \item We then simplified the prompt but edited examples to avoid phrases like ``images of...'', resulting in captions more compatible with similarity models.
\end{itemize}

\paragraph{LLM Comparison for Caption Generation.} We compared three LLMs: \texttt{mistralai/Mistral-7B-Instruct-v0.2}, \texttt{Qwen/Qwen2.5-14B-Instruct}, and \texttt{Qwen/Qwen2.5-32B-Instruct}. The 32B model produced the most diverse and generalizable difference captions, while smaller models tended to overfit or focus narrowly on individual images.

\paragraph{Summarizing Conceptual Differences.} Finally, we prompted LLMs to summarize conceptual shifts based on similarity deltas and difference captions. Larger models generated the most natural and high-level descriptions; \texttt{Qwen2.5-32B-Instruct} performed reasonably well.

Overall, while these results show some promise (Table~\ref{tab:caption_diffs}) for settings with images with common semantic properties, they are relatively inconsistent. Moreover, for datasets without such common semantic properties, e.g., medical images, these methods may only work with dedicated foundation models appropriate for those image modalities.

\begin{figure}[h]
\begin{tcolorbox}[colback=white, colframe=mitred,
  title=Caption Prompt (Terra Incognita),
  fonttitle=\bfseries,
  coltitle=white,
  boxrule=0.8pt,
  enhanced,
  breakable]
\tiny
Caption this image. I know what object it is. Focus on describing the artistic style, texture, and domain-specific details rather than the object itself. Again, do not mention the object class.
\end{tcolorbox}

\begin{tcolorbox}[colback=white, colframe=mitred,
  title=Differences Prompt (TerraIncognita),
  fonttitle=\bfseries,
  coltitle=white,
  boxrule=0.8pt,
  enhanced,
  breakable]

\tiny
I am a machine learning researcher trying to figure out properties of an image beyond the image class.             Give me a description of this image that is more specific than the image class. For instance,             if this is an image of a bird, I don't want to know if the bird is a sparrow or a crow. I want to know             if the bird is flying or sitting on a branch, if the camera is a drone or a ground-level camera,             if the image is a macro shot or a close-up, or if the lighting is natural or artificial.             A broader list of such properties is what I'm looking for. Give me this description for the image without focusing on the animal class.\\

Come up with \texttt{<number of captions>} distinct concepts that are more likely to be true for the Out-of-Distribution Group compared to the In-Distribution Group. Please write a list of captions (separated by bullet points \texttt{"*"}). For example:
\begin{itemize}
  \item \texttt{"unusual lighting conditions"}
  \item \texttt{"visual distortions"}
  \item \texttt{"complex backgrounds"}
  \item \texttt{"non-standard object poses"}
  \item \texttt{"uncommon viewing angles"}
  \item \texttt{"partial views of objects"}
  \item \texttt{"objects in unexpected contexts"}
  \item \texttt{"scenes with high visual clutter"}
  \item \texttt{"images with unusual color schemes"}
  \item \texttt{"low-resolution images"}
\end{itemize}

Do not talk about the caption itself, e.g., \texttt{"caption with one word"}, and do not list more than one concept. The hypothesis should be a caption, so phrasing like \texttt{"more of ..."}, \texttt{"presence of ..."}, or \texttt{"images with ..."} is incorrect. Also, do not enumerate possibilities within parentheses. Here are examples of bad outputs and their corrections:
\begin{itemize}
  \item INCORRECT: \texttt{"various nature environments like lakes, forests, and mountains"} \quad CORRECTED: \texttt{"nature"}
  \item INCORRECT: \texttt{"images of household object (e.g. bowl, vacuum, lamp)"} \quad CORRECTED: \texttt{"household objects"}
  \item INCORRECT: \texttt{"Presence of baby animals"} \quad CORRECTED: \texttt{"baby animals"}
  \item INCORRECT: \texttt{"Different types of vehicles including cars, trucks, boats, and RVs"} \quad CORRECTED: \texttt{"vehicles"}
  \item INCORRECT: \texttt{"Images involving interaction between humans and animals"} \quad CORRECTED: \texttt{"interaction between humans and animals"}
  \item INCORRECT: \texttt{"More realistic images"} \quad CORRECTED: \texttt{"realistic images"}
  \item INCORRECT: \texttt{"Insects (cockroach, dragonfly, grasshopper)"} \quad CORRECTED: \texttt{"insects"}
\end{itemize}

Again, I want to figure out what kind of distribution shift there is. List \texttt{<number of captions>} properties that hold more often for the images (not captions) in the Out-of-Distribution Group compared to the In-Distribution Group. Answer with a list (separated by bullet points \texttt{"*"}).\\

\textbf{In-Distribution Group:} \texttt{<list of in-distribution captions>}\\

\textbf{Out-of-Distribution Group:} \texttt{<list of out-of-distribution captions>}\\

\textbf{Your response:}
\end{tcolorbox}
\caption{ID/OOD difference prompt for TerraIncognita.}
\end{figure}

\begin{figure}[h]
\begin{tcolorbox}[colback=white, colframe=mitred,
  title=Differences Prompt (PACS),
  fonttitle=\bfseries,
  coltitle=white,
  boxrule=0.8pt,
  enhanced,
  breakable]
\tiny
Caption this image. I know what object it is. Focus on describing contextual and environmental details, such as scene composition, lighting, and background characteristics. Again, do not mention the object class.
\end{tcolorbox}

\begin{tcolorbox}[colback=white, colframe=mitred,
  title=Differences Prompt (PACS),
  fonttitle=\bfseries,
  coltitle=white,
  boxrule=0.8pt,
  enhanced,
  breakable]

\tiny
I am a machine learning researcher trying to figure out properties of an image beyond the object class.             Give me a description of this image that is more specific than the object class. For instance,             if this is an image of a dog, I don't want to know if the dog is a bulldog or a retriever. I want to know             if the scene suggests an indoor or outdoor setting, details about the artistic style, or specific texture and lighting.             A broader list of such properties is what I'm looking for. Give me this description for the image without mentioning the object class.\\

Come up with \texttt{<number of captions>} distinct concepts that are more likely to be true for the Out-of-Distribution Group compared to the In-Distribution Group. Please write a list of captions (separated by bullet points \texttt{"*"}). For example:
\begin{itemize}
  \item \texttt{"unusual lighting conditions"}
  \item \texttt{"visual distortions"}
  \item \texttt{"complex backgrounds"}
  \item \texttt{"non-standard object poses"}
  \item \texttt{"uncommon viewing angles"}
  \item \texttt{"partial views of objects"}
  \item \texttt{"objects in unexpected contexts"}
  \item \texttt{"scenes with high visual clutter"}
  \item \texttt{"images with unusual color schemes"}i
  \item \texttt{"low-resolution images"}
\end{itemize}

Do not talk about the caption itself, e.g., \texttt{"caption with one word"}, and do not list more than one concept. The hypothesis should be a caption, so phrasing like \texttt{"more of ..."}, \texttt{"presence of ..."}, or \texttt{"images with ..."} is incorrect. Also, do not enumerate possibilities within parentheses. Here are examples of bad outputs and their corrections:
\begin{itemize}
  \item INCORRECT: \texttt{"various nature environments like lakes, forests, and mountains"} \quad CORRECTED: \texttt{"nature"}
  \item INCORRECT: \texttt{"images of household object (e.g. bowl, vacuum, lamp)"} \quad CORRECTED: \texttt{"household objects"}
  \item INCORRECT: \texttt{"Presence of baby animals"} \quad CORRECTED: \texttt{"baby animals"}
  \item INCORRECT: \texttt{"Different types of vehicles including cars, trucks, boats, and RVs"} \quad CORRECTED: \texttt{"vehicles"}
  \item INCORRECT: \texttt{"Images involving interaction between humans and animals"} \quad CORRECTED: \texttt{"interaction between humans and animals"}
  \item INCORRECT: \texttt{"More realistic images"} \quad CORRECTED: \texttt{"realistic images"}
  \item INCORRECT: \texttt{"Insects (cockroach, dragonfly, grasshopper)"} \quad CORRECTED: \texttt{"insects"}
\end{itemize}

Again, I want to figure out what kind of distribution shift there is. List \texttt{<number of captions>} properties that hold more often for the images (not captions) in the Out-of-Distribution Group compared to the In-Distribution Group. Answer with a list (separated by bullet points \texttt{"*"}).\\

\textbf{In-Distribution Group:} \texttt{<list of in-distribution captions>}\\

\textbf{Out-of-Distribution Group:} \texttt{<list of out-of-distribution captions>}\\

\textbf{Your response:}
\end{tcolorbox}
\caption{ID/OOD difference prompt for PACS.}
\end{figure}

\begin{figure}[h]
\begin{tcolorbox}[colback=white, colframe=mitred,
  title=Caption Prompt (VLCS),
  fonttitle=\bfseries,
  coltitle=white,
  boxrule=0.8pt,
  enhanced,
  breakable]
\tiny
Caption this histopathology image. I know its diagnostic category. Focus on describing tissue morphology, staining patterns, and structural details. Again, do not reveal the diagnostic category.
\end{tcolorbox}

\begin{tcolorbox}[colback=white, colframe=mitred,
  title=Differences Prompt (VLCS),
  fonttitle=\bfseries,
  coltitle=white,
  boxrule=0.8pt,
  enhanced,
  breakable]

\tiny
I am a researcher studying domain adaptation. Please describe this image with a focus on properties beyond the object class.             For example, if this is an image of a bird, I don't want to know whether it is a sparrow or an eagle.             Instead, I want detailed information about the environmental context, scene composition, lighting conditions, and background.             Provide such a description without mentioning the object class.\\

Come up with \texttt{<number of captions>} distinct concepts that are more likely to be true for the Out-of-Distribution Group compared to the In-Distribution Group. Please write a list of captions (separated by bullet points \texttt{"*"}). For example:
\begin{itemize}
  \item \texttt{"unusual lighting conditions"}
  \item \texttt{"visual distortions"}
  \item \texttt{"complex backgrounds"}
  \item \texttt{"non-standard object poses"}
  \item \texttt{"uncommon viewing angles"}
  \item \texttt{"partial views of objects"}
  \item \texttt{"objects in unexpected contexts"}
  \item \texttt{"scenes with high visual clutter"}
  \item \texttt{"images with unusual color schemes"}
  \item \texttt{"low-resolution images"}
\end{itemize}

Do not talk about the caption itself, e.g., \texttt{"caption with one word"}, and do not list more than one concept. The hypothesis should be a caption, so phrasing like \texttt{"more of ..."}, \texttt{"presence of ..."}, or \texttt{"images with ..."} is incorrect. Also, do not enumerate possibilities within parentheses. Here are examples of bad outputs and their corrections:
\begin{itemize}
  \item INCORRECT: \texttt{"various nature environments like lakes, forests, and mountains"} \quad CORRECTED: \texttt{"nature"}
  \item INCORRECT: \texttt{"images of household object (e.g. bowl, vacuum, lamp)"} \quad CORRECTED: \texttt{"household objects"}
  \item INCORRECT: \texttt{"Presence of baby animals"} \quad CORRECTED: \texttt{"baby animals"}
  \item INCORRECT: \texttt{"Different types of vehicles including cars, trucks, boats, and RVs"} \quad CORRECTED: \texttt{"vehicles"}
  \item INCORRECT: \texttt{"Images involving interaction between humans and animals"} \quad CORRECTED: \texttt{"interaction between humans and animals"}
  \item INCORRECT: \texttt{"More realistic images"} \quad CORRECTED: \texttt{"realistic images"}
  \item INCORRECT: \texttt{"Insects (cockroach, dragonfly, grasshopper)"} \quad CORRECTED: \texttt{"insects"}
\end{itemize}

Again, I want to figure out what kind of distribution shift there is. List \texttt{<number of captions>} properties that hold more often for the images (not captions) in the Out-of-Distribution Group compared to the In-Distribution Group. Answer with a list (separated by bullet points \texttt{"*"}).\\

\textbf{In-Distribution Group:} \texttt{<list of in-distribution captions>}\\

\textbf{Out-of-Distribution Group:} \texttt{<list of out-of-distribution captions>}\\

\textbf{Your response:}
\end{tcolorbox}
\caption{ID/OOD difference prompt for VLCS.}
\end{figure}

\FloatBarrier
\end{document}